\documentclass[letterpaper, 10 pt, conference]{ieeeconf}  % Comment this line out if you need a4paper

\IEEEoverridecommandlockouts                              % This command is only needed if 
\usepackage[style=ieee, url=false, doi=false, natbib=true, mincitenames=1, maxcitenames=1]{biblatex}

\usepackage{hyperref}

\usepackage{tikz}
\usepackage{mwe}
\usetikzlibrary{arrows,decorations,backgrounds,calc}
\usetikzlibrary{positioning, arrows.meta}
\usepackage{relsize}
\usepackage{pgfplots}
\usepgfplotslibrary{groupplots}
\pgfplotsset{compat=1.14}
\usepackage{booktabs}
\usepackage{color}
\usepackage{subcaption}
\usepackage{wrapfig}
\usepackage{graphicx,caption}
\usepackage{amsmath}
\usepackage{mathtools}
\usepackage{bbm}
\usepackage{bm}
\usepackage{siunitx}
\usepackage{amsfonts}
\usepackage[capitalise]{cleveref}

\DeclarePairedDelimiterX{\infdivx}[2]{(}{)}{%
  #1\;\delimsize|\delimsize|\;#2%
}

\newcommand{\kld}[2]{\ensuremath{\text{KL}\infdivx{#1}{#2}}}

\title{\LARGE \bf
Multi-Agent Variational Occlusion Inference\\Using People as Sensors
}

\author{ Masha Itkina$^{1}$, Ye-Ji Mun$^{2}$, Katherine Driggs-Campbell$^{2}$, and Mykel J. Kochenderfer$^{1}$% <-this % stops a space
\thanks{This project was supported by funding from the Ford-Stanford Alliance and a gift from Mercedes-Benz Research \& Development North America.}
\thanks{We thank Spencer M. Richards and Ransalu Senanayake for their invaluable feedback.}
\thanks{$^{1}$Masha Itkina and Mykel J. Kochenderfer are with the Aeronautics and Astronautics Department, Stanford University, USA. Email: {\tt \{mitkina, mykel\}@stanford.edu}.}%
\thanks{$^{2}$Ye-Ji Mun and Katherine Driggs-Campbell are with the Electrical and Computer Engineering
Department, University of Illinois at Urbana-Champaign, USA. Email: {\tt \{yejimun2, krdc\}@illinois.edu}.}%
}

\begin{document}

\maketitle
\thispagestyle{empty}
\pagestyle{empty}

%%%%%%%%%%%%%%%%%%%%%%%%%%%%%%%%%%%%%%%%%%%%%%%%%%%%%%%%%%%%%%%%%%%%%%%%%%%%%%%%
\begin{abstract}
Autonomous vehicles must reason about spatial occlusions in urban environments to ensure safety without being overly cautious. Prior work explored occlusion inference from observed social behaviors of road agents, hence treating \textit{people as sensors}. Inferring occupancy from agent behaviors is an inherently multimodal problem; a driver may behave similarly for different occupancy patterns ahead of them (e.g., a driver may move at constant speed in traffic or on an open road). Past work, however, does not account for this multimodality, thus neglecting to model this source of aleatoric uncertainty in the relationship between driver behaviors and their environment. We propose an occlusion inference method that characterizes observed behaviors of human agents as sensor measurements, and fuses them with those from a standard sensor suite. To capture the aleatoric uncertainty, we train a conditional variational autoencoder with a discrete latent space to learn a multimodal mapping from observed driver trajectories to an occupancy grid representation of the view ahead of the driver. Our method handles multi-agent scenarios, combining measurements from multiple observed drivers using evidential theory to solve the sensor fusion problem. Our approach is validated on a cluttered, real-world intersection, outperforming baselines and demonstrating real-time capable performance. Our code is available at \href{https://github.com/sisl/MultiAgentVariationalOcclusionInference}{https://github.com/sisl/MultiAgentVariationalOcclusionInference}.

\end{abstract}

\section{Introduction}
\label{sec:intro}
Safe autonomous navigation in the presence of occlusions in a cluttered, human environment is an open problem for mobile robot perception and decision making~\cite{Bouton2018aamas}. It is common in robotics to assume occluded regions to be either free~\cite{koenig2002d} or occupied space~\cite{florence2020integrated}, which may result in either dangerous or overly cautious behavior. While humans suffer from similar limitations, they are able to make inferences about these unobserved spaces using insights from prior experience as well as semantic and geometric information. While driving, humans intuitively anticipate potential hazards, even when they are occluded, by observing the behavior of other road agents. For instance, if another vehicle brakes sharply, a driver may infer the presence of an occluded obstacle (e.g., a pedestrian) ahead, necessitating them to act cautiously. Likewise, autonomous vehicles should be capable of making inferences about occluded regions in order to safely interact with human road users. 

Occlusion inference in the context of interaction has been explored from the perspectives of mapping~\cite{people_as_sensors}, inverse reinforcement learning~\cite{socialperception2019}, and pedestrian tracking~\cite{hara2020pedestrians}. \citet{people_as_sensors} used driver models as additional sensor information to provide ``measurements'' within regions of occlusion in an environment map, coining the term \emph{People as Sensors (PaS)}. They showed occupancy grid maps (OGMs)~\cite{occupancy_grids} are a useful environment representation for fusing insights from observed interactive behaviors with traditional sensor outputs. The OGM representation does not require prior knowledge of the environment structure and can handle partially observed and previously unseen road users, making it more generalizable compared to object-centric alternatives~\cite{itkina2019dynamic}.

Occlusion inference based on observed behaviors is an intrinsically multimodal problem. For example, a driver may drive at the speed limit in traffic (occupied space ahead) or on an empty road (free space ahead). Similarly, a driver may brake for a single pedestrian or a group of pedestrians, which would yield different occupancy patterns. The occlusion inference model must handle this irreducible (i.e., aleatoric) uncertainty to accurately represent the space of possible inferred maps. However, prior work~\cite{people_as_sensors, socialperception2019, hara2020pedestrians} often does not account for the multimodality in the occluded region's
\begin{figure}[t!]
    \centering
    \scalebox{0.75}{\input{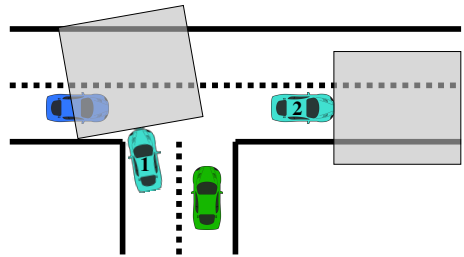}}
    \caption{ %
    Our proposed occlusion inference approach. The learned driver sensor model maps behaviors of visible drivers (cyan) to an OGM of the environment ahead of them (gray). The inferred OGMs are then fused into the ego vehicle's (green) map. Our goal is to infer the presence or absence of occluded agents (blue). In this scenario, driver 1 is waiting to turn left, occluding oncoming traffic from the ego vehicle's view. Driver 1 being stopped should indicate to the ego vehicle that there may be oncoming traffic, and thus it is not safe to proceed with its right turn. Driver 2 is driving at a constant speed. This observed behavior is not sufficient to discern whether the vehicle is traveling in traffic or on an open road. 
    We aim to encode such intuition into our occlusion inference algorithm.
    }
    \label{fig:approach}
    \vspace{-5pt}
\end{figure}
occupancy given a driver behavior, thus neglecting this source of aleatoric uncertainty.

Inspired by PaS~\cite{people_as_sensors}, we use observed road agent behaviors as additional sensor information into an OGM of the environment. We extend PaS to model the multimodality in the distribution and scale it to a multi-agent framework where several human drivers can provide measurements within the occluded regions of the map. To do this, we present an occlusion inference algorithm with two stages. First, we learn a \emph{driver sensor model} that maps an observed driver trajectory to a discrete set of possibilities for the OGM of the space ahead of the driver. To account for the multimodality in the inferred occupancy, we use a conditional variational autoencoder (CVAE)~\cite{cvae} with a discrete latent space for the driver sensor model. In the second stage, we fuse the spatial predictions inferred from multiple observed driver trajectories into the environment map using a multi-agent sensor fusion mechanism based in evidential theory~\cite{dst}. 

The goal of our approach is to enhance current perception and inference techniques specifically targeting unmeasured, occluded spaces by combining ideas from mapping, human modeling, and sensor fusion. We present a framework that can handle complex multi-agent settings and validate the proposed technique using real-world data. Our contributions are as follows: 
1) We learn a sensor model for human drivers end-to-end using a CVAE architecture with a discrete latent space that captures multimodality in the distribution;
2) Our occlusion inference method allows us to consider multiple human drivers (modeled as sensors) through evidential sensor fusion; and
3) We demonstrate that our method is real-time capable and outperforms baselines on a cluttered, unsignalized intersection in the INTERACTION dataset~\cite{interaction}.
\section{Related Works}
\label{sec:related_works}

\noindent\textbf{Temporary Occlusions.}
Numerous methods have explored tracking objects through temporary occlusions (i.e., an object coming in and out of view). 
In robot manipulation, a self-occlusion results when the robot temporarily occludes objects of interest. This temporary issue is typically dealt with using specialized architectures and features~\cite{ebert2017self, wang2020roll,Park-RSS-20}. These works usually assume a static background, which is not the case in dynamic urban driving. \citet{dequaire2018deep} handle temporary occlusions in an urban environment by learning to hallucinate previously observed objects in occluded regions from memory. Our method differs from memory-based occlusion inference by considering interactive behavior as an additional information source to infer fully occluded, previously unobserved obstacles in a self-supervised manner.

\noindent\textbf{Inpainting Urban Scenes.}
A separate line of work from computer vision uses inpainting of the environment to fill in the occluded portions of the scene. 
Inpainting has been used to infer bird's-eye view (BeV) semantics of the partially occluded static environment behind foreground obstacles~\cite{schulter2018learning, lu2020semantic} as well as the partially occluded foreground obstacles~\cite{purkait2019seeing}. Inpainting of the BeV environment semantics provides a more comprehensive estimate of the navigable space and can improve path planning as compared to the original more restricted, occluded view~\cite{han2020planning}. In occupancy mapping, Gaussian processes have been used to interpolate occupancy behind partially occluded obstacles~\cite{gpomsramos2012,senanayake2017corl,senanayake2018corl}. Most of these methods assume static environments, and none of them reason about fully occluded, dynamic obstacles.

\noindent\textbf{Occlusion Inference Using Environment Structure.} 
Other works infer fully occluded road agents using prior knowledge of behaviors, dynamics, and environment structure. These approaches use this information to define reachable regions for potentially occluded obstacles~\cite{frazzoli2019, hubmann2019pomdp}. Hypothesized road agents can be placed at the limits of the observable space~\cite{chae2020virtual} or in occluded road locations that might lead to collisions with the ego vehicle~\cite{hubmann2019pomdp}. Their velocity profiles can be predetermined by their class~\cite{frazzoli2019} or using information from a digital map~\cite{hoermann_occlusion}. These approaches often do not exploit observable interactions and only consider worst-case scenarios which leads to overly conservative behavior.

\noindent\textbf{Social Occlusion Inference.}
Methods that use observed social interactions to infer fully occluded obstacles are most closely related to our work.
\citet{hara2020pedestrians} infer the existence of an approaching car outside the field of view using observed pedestrian reactions. \citet{socialperception2019} perform behavior-based occlusion inference in simulation by learning a cost function that acts as the driver sensor model. These costs update the belief over visible object states and hidden states, such as driving styles. \citet{people_as_sensors} uncover driver actions by clustering over trajectories and then learn occupancy probabilities for the OGM for each action. This two-step process forms their driver sensor model. The formulation is demonstrated on a crosswalk scenario with a single driver sensor and a single occluded pedestrian. We propose an end-to-end approach, which learns a multimodal mapping from driver trajectories to the OGM ahead of the driver, thereby accounting for aleatoric uncertainty. We generalize the technique to multi-agent settings using ideas from sensor fusion and validate our method on real-world data.
\section{Problem Formulation}
\label{sec:problem_formulation}
We represent the ego vehicle's local environment as an OGM $\mathcal{M}_{ego}^{obs} \in [0, 1]^{H \times W}$, where $H$ and $W$ are its dimensions. The OGM is generated using ray tracing from a range-bearing sensor in the standard sensor suite (e.g., LiDAR or radar). The occlusion inference task then becomes imputing occupancy probabilities in regions of occlusion within this OGM. Occluded regions refer to the subset of cells in the OGM that receive no sensor measurements. Assuming accurate sensors, the observed occupied grid cells have an occupancy probability of $1$, while free cells have an occupancy probability of $0$. Occluded grid cells are denoted with $0.5$ occupancy probability. Without loss of generality, we further assume that the OGMs do not incorporate temporal history. The omniscient OGM $\mathcal{M}_{ego}^{gt} \in \{0, 1\}^{H \times W}$ contains the ground truth occupancy from a BeV without occlusions. We aim to infer the occupancy of the occluded grid cells (where $\mathcal{M}_{ego}^{obs} = 0.5$) from observed human driver behaviors.

We assume that our ego vehicle is equipped with a perception system that is able to detect, track, and estimate the pose of visible road agents, which is a reasonable assumption given existing computer vision capabilities~\cite{interaction,caesar2020nuscenes}. We denote the state of an observed human driver $h$ at time $t$ as,%
\begin{equation}\label{eq:vec}
    s_{h}^{t} = \left[x_{h}^{t}, y_{h}^{t}, \psi_{h}^{t}, v_{x, h}^{t}, v_{y, h}^{t}, a_{x, h}^{t}, a_{y, h}^{t} \right],
\end{equation}
\noindent capturing the position, orientation, velocity, and acceleration. Each human driver $h$ observes the environment ahead of them represented as another OGM $\mathcal{M}_{h}~\in~\{0, 1\}^{H_{h}~\times W_{h}}$, where $H_{h}$ and $W_{h}$ denote the dimensions of the human driver's OGM. As visualized in \cref{fig:approach}, the view represented by $\mathcal{M}_{h}$ may be occluded from the ego vehicle. We hypothesize that the region ahead of the human driver may be partially recovered from their observed behavior. Thus, we use the observed behaviors and interactions of human drivers to impute occupancy in occluded regions of $\mathcal{M}_{ego}^{obs}$.
\begin{figure}[t!]
    \centering
    \vspace{3.5pt}
    \scalebox{0.65}{\input{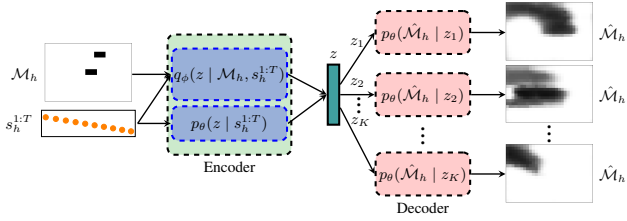}}
    \caption{The proposed driver sensor model is a CVAE with two encoders and one decoder. The encoders learn the categorical prior and posterior distributions $p_{\theta}(z \!\mid\! s_{h}^{1:T})$ and $q_{\phi}(z \!\mid\! s_{h}^{1:T}, \mathcal{M}_{h})$, respectively, over the discrete latent variable $z \in \{1, \hdots, K\}$, where $s_{h}^{1:T}$ is the observed trajectory of driver $h$, $\mathcal{M}_{h}$ is the ground truth OGM of the view ahead of driver $h$, and $\theta$ and $\phi$ are network parameters. 
     The decoder maps each latent class $z$ to an estimated environment view $\hat{\mathcal{M}}_{h}$ ahead of the driver. During inference, given an input trajectory $s_{h}^{1:T}$, the network produces a multimodal categorical prior over the possible OGMs $\hat{\mathcal{M}}_{h}$ corresponding to each latent class $z$.}
    \label{fig:architecture}
    \vspace{-10pt}
\end{figure}
\section{Multi-Agent Variational Occlusion Inference}
\label{sec:methods}
Our proposed occlusion inference approach, illustrated in \cref{fig:approach}, consists of two steps. First, we learn a driver sensor model that maps an observed trajectory $s^{1:T}_{h}$ of a driver~$h$ over $T$ time steps to the OGM representation $\mathcal{M}_{h}$ of the environment ahead of them at time $T$. Second, the inferred OGMs $\mathcal{M}_{h}$ for all observed drivers $h$ are fused into $\mathcal{M}_{ego}^{obs}$ in regions of occlusion as additional sensor measurements. 

\noindent\textbf{Driver Sensor Model.}
We draw on human behavior modeling literature~\cite{brown2020taxonomy} to build our driver sensor model. A common approach in interaction modeling is to discretize the set of human behaviors~\cite{kotseruba2016joint,bandyopadhyay2013intention}. CVAEs have been shown to successfully discretize such behaviors and capture their aleatoric uncertainty~\cite{schmerling2018multimodal, salzmann2020trajectron}. As such, we formulate our driver sensor model as a CVAE that maps observed trajectories to a discrete latent space that captures the multimodal possibilities for the environment ahead of the driver.

The proposed CVAE, illustrated in \cref{fig:architecture}, consists of two encoders and a decoder, as inspired by the setup of~\citet{itkina2020evidential}. The encoders represent the categorical prior and posterior distributions $p_{\theta}(z~\mid s_{h}^{1:T})$ and $q_{\phi}(z \!\mid\!s_{h}^{1:T}, \mathcal{M}_{h})$, respectively, over the discrete latent variable $z \in \{1, \hdots, K\}$, where $\theta$ and $\phi$ are network parameters. The latent encoding~$z$ is distributed according to the posterior during training and the prior during inference. The decoder uses $z$ to output an estimated environment view $\hat{\mathcal{M}}_{h} \in [0,1]^{H_{h}~\times W_{h}}$ for the driver. The learned categorical prior $p_{\theta}(z \!\mid\! s_{h}^{1:T})$ accounts for the multimodal distribution over the inferred environment. During inference, the network maps the trajectory $s_{h}^{1:T}$ to the categorical prior over the possible OGMs $\hat{\mathcal{M}}_{h}$ corresponding to each latent class $z \in \{1, \hdots, K\}$.

We train the CVAE with a modified evidence lower bound (ELBO) loss, 
\begin{equation} \label{eq:loss}
\begin{aligned}
    \mathcal{L} (\mathcal{M}_{h}, s^{1:T}_{h}; \theta, \phi) = &-\mathbb{E}_{z \sim q_{\phi}(z \mid s_{h}^{1:T})} [\log p_{\theta}(\mathcal{M}_{h}|z)] \\ &+ \beta \kld{q_{\phi}(z \!\mid\! s^{1:T}_{h}, \mathcal{M}_{h})}{p_{\theta}(z \!\mid\! s^{1:T}_{h})}\\
    &- \alpha I_{q_{\phi}}(\mathcal{M}_{h}; z),
\end{aligned}
\end{equation}
where $\alpha$ and $\beta$ are hyperparameter weights, $\text{KL}$ is the Kullback–Leibler divergence, and $I_{q_{\phi}}(\mathcal{M}_{h}; z)$ is the approximate mutual information between $\mathcal{M}_{h}$ and $z$ under the posterior distribution ${q_{\phi}(z \!\mid\! s_{h}^{1:T}, \mathcal{M}_{h})}$~\cite{salzmann2020trajectron}. 
% We use the prior ${p_{\theta}(z \!\mid\! s_{h}^{1:T})}$ as a proxy for the posterior distribution and acquire $p_{\theta}(z)$ by summing over the batch following~\cite{salzmann2020trajectron}. 
By keeping the number of latent classes $K$ small, we are able to compute the reconstruction loss term (first term in \cref{eq:loss}) exactly without resorting to sampling approximation techniques such as the Gumbel-Softmax reparameterization~\cite{gumbel_softmax, concrete}. To address the class imbalance between occupied and free cells in the data, the reconstruction loss is weighted according to one minus the fraction of occupied or free cells in a batch. The full loss is computed as the average of \cref{eq:loss} over a batch.

\noindent\textbf{Multi-Sensor Fusion for Occlusion Inference.}
The driver sensor model outputs $K$ estimated OGMs $\hat{\mathcal{M}}_{h}$ and the associated learned categorical prior $p_{\theta}(z \!\mid\! s_{h}^{1:T})$ given an observed trajectory $s_{h}^{1:T}$ for driver $h$. These OGMs for all visible drivers~$h$ need to be incorporated into $\mathcal{M}_{ego}^{obs}$. The occlusion inference problem then becomes a sensor fusion task. We use evidential theory~\cite{dst}, a popular sensor fusion technique~\cite{kleinsensorfusion}, to fuse the measurements from the driver sensor model into $\mathcal{M}_{ego}^{obs}$. Evidential theory is able to discern conflicting information (e.g., two differing sensor measurements) from lack of information (e.g., a region of occlusion) by considering the power set of exhaustive hypotheses. For OGMs, our hypothesis set is $\Omega = \{O, F\}$, where $O$ is occupied and $F$ is free. The power set of the hypotheses is then $2^{\Omega}~=~\{\emptyset, \{O\},~\{F\},~\{O, F\}\}$. Evidential theory defines a belief mass function $m~:~2^{\Omega}~\to~[0,1]$ such that $\sum_{A \subseteq \Omega} m(A)~=~1$. Since a grid cell must be either free or occupied, the belief mass for the empty set is zero: $m(\emptyset) = 0$. A non-zero belief mass on set $\{O, F\}$ represents a lack of evidence that discerns between the occupied and free hypotheses. Belief mass functions from different sensors can be combined using Dempster-Shafer's rule~\cite{dst}.

We begin the sensor fusion process by heuristically transforming the probabilities in $\hat{\mathcal{M}}_{h}$ into belief masses for each observed human driver $h$ in the vicinity of the ego vehicle:
\begin{equation}
    m_{h,c}(\{O\}) = \delta\hat{\mathcal{M}}_{h,c}, \quad m_{h,c}(\{F\}) = \delta(1-\hat{\mathcal{M}}_{h,c}),
\end{equation}
\noindent where $c$ is a grid cell and $\delta \in [0,1]$ is a hyperparameter. Intuitively, the occupancy probability in $\hat{\mathcal{M}}_{h,c}$ is considered evidence towards the occupied class and, symmetrically, the free probability $1 - \hat{\mathcal{M}}_{h,c}$ is considered evidence towards the free class. Both are scaled down by a factor $\delta$, which represents the uncertainty belief mass $m_{h,c}(\{O,F\}) = 1-\delta$. We set $\delta$ to 0.95 in our experiments.

We incorporate the inferred occupancy estimates only in the occluded cells of $\mathcal{M}_{ego}^{obs}$ since the driver sensor model will be less precise than traditional sensors (e.g., LiDAR) in unoccluded portions of the map. Prior to receiving any measurements in an occluded cell $c$, we have complete uncertainty in its occupancy, i.e., $m_{ego,c}(\{O,F\}) = 1$. We sequentially update this belief mass with the measurement masses $m_{h,c}$ of the observed drivers for a single time step using Dempster-Shafer's rule~\cite{dst}:
\begin{equation}
\label{eq:dst_rule}
\begin{aligned}
    m_{ego,c}\left(A\right) 
    &= m_{ego,c} \oplus m_{h,c}\left(A\right)
    \\ 
    &\coloneqq \frac{\sum_{X \cap Y = A}m_{ego,c}\left(X\right)m_{h,c}\left(Y\right)}{1 - \sum_{X \cap Y = \emptyset} m_{ego,c}\left(X\right)m_{h,c}\left(Y\right)},
\end{aligned}
\end{equation}
for all $A, X, Y \in 2^{\Omega}$. The masses $m_{ego,c}$ are converted to traditional probabilities using the pignistic transform~\cite{pignistic}:%
\begin{equation} \label{eq:pignistic}
     \hat{\mathcal{M}}_{ego,c}\left(B\right) = \sum_{A \in 2^{\Omega}} m_{ego,c}\left(A\right) \frac{\vert B \cap A \vert}{\vert A \vert},
\end{equation}
where $B$ is a singleton hypothesis and $\vert \cdot \vert$ is the cardinality of a set.%

We compute the likelihood of a fused grid $\hat{\mathcal{M}}_{ego}$ using the learned priors $p_{\theta}(z \!\mid\! s_{h}^{1:T})$ from the driver sensor model. We assume the trajectory $s_{h}^{1:T}$ and the latent encoding~$z$ data pairs are 
i.i.d.. Then, the likelihood of $\hat{\mathcal{M}}_{ego}$ is $p(\hat{\mathcal{M}}_{ego})~=~\prod_{h} p_{\theta}(z \!\mid\! s_{h}^{1:T})$ for all visible drivers~$h$.
\begin{table*}[t!]
\small
\begin{center}
\vspace{3pt}
\caption{\footnotesize Our driver sensor model and full occlusion inference pipeline consistently outperform baselines across metrics on the test set. Bold denotes the best performing model across a metric. Note that IS values are divided by 100. For the driver sensor model (full occlusion pipeline), the maximum standard error per grid cell for accuracy and MSE is 0.0007 (0.0008), and per OGM for IS is 0.0022 (0.0046). 
} \label{tab:sensor_model}

\begin{tabular}{@{}l|rrr|rrr|rrr|rrr@{}}
\hline
& \multicolumn{6}{c|}{Driver Sensor Model} & \multicolumn{6}{c}{Full Occlusion Inference Pipeline}\\
\hline\hline
& Occ. & Free & Overall & Occ. & Free & Overall & Occ. & Free & Overall & Occ. & Free & Overall\\
\hline
Method & \multicolumn{3}{c|}{Acc. $\uparrow$} &
\multicolumn{3}{c|}{Top 3 Acc. $\uparrow$} & \multicolumn{3}{c|}{Acc. $\uparrow$} &
\multicolumn{3}{c}{Top 3 Acc. $\uparrow$}\\
\hline
K-means PaS~\cite{people_as_sensors} & 0.512 & 0.463 & 0.465 & N/A & N/A & N/A & 0.834 & 0.680 & 0.682 & N/A & N/A & N/A\\
GMM PaS & 0.494 & 0.439 & 0.440 & 0.611 & 0.562 & 0.557 & \textbf{0.838} & 0.660 & 0.663 & \textbf{0.860} & 0.691 & 0.691\\
Ours & \textbf{0.619} & \textbf{0.601} & \textbf{0.601} & \textbf{0.819} & \textbf{0.773} & \textbf{0.764} & 0.660 & \textbf{0.722} & \textbf{0.722} & 0.746 & \textbf{0.778} & \textbf{0.774}\\
\hline
& \multicolumn{3}{c|}{MSE $\downarrow$} &
\multicolumn{3}{c|}{Top 3 MSE $\downarrow$} & \multicolumn{3}{c|}{MSE $\downarrow$} &
\multicolumn{3}{c}{Top 3 MSE $\downarrow$}\\
\hline
K-means PaS~\cite{people_as_sensors} & \textbf{0.188} & 0.206 & 0.206 & N/A & N/A & N/A & 0.146 & 0.198 & 0.197 & N/A & N/A & N/A\\
GMM PaS & 0.192 & 0.211 & 0.211 & 0.157 & 0.166 & 0.169 & \textbf{0.144} & 0.205 & 0.204 & \textbf{0.123} & 0.181 & 0.181\\
Ours & 0.293 & \textbf{0.192} & \textbf{0.194} & \textbf{0.145} & \textbf{0.113} & \textbf{0.124} & 0.303 & \textbf{0.171} & \textbf{0.173} & 0.233 & \textbf{0.136} & \textbf{0.140}\\
\hline
& \multicolumn{3}{c|}{IS $\downarrow$} &
\multicolumn{3}{c|}{Top 3 IS $\downarrow$} & \multicolumn{3}{c|}{IS $\downarrow$} &
\multicolumn{3}{c}{Top 3 IS $\downarrow$}\\
\hline
K-means PaS~\cite{people_as_sensors} & 0.205 & 0.020 & 0.225 & N/A & N/A & N/A & 1.373 & 0.027 & 1.400 & N/A & N/A & N/A\\
GMM PaS & 0.209 & 0.026 & 0.235 & 0.185 & 0.015 & 0.209 & 1.393 & 0.029 & 1.423 & 1.277 & 0.018 & 1.297\\
Ours & \textbf{0.187} & \textbf{0.017} & \textbf{0.204} & \textbf{0.115} & \textbf{0.007} & \textbf{0.130} & \textbf{1.336} & \textbf{0.017} & \textbf{1.353} & \textbf{1.220} & \textbf{0.011} & \textbf{1.232}\\
\hline
\end{tabular}
\end{center}
\vspace{-20pt}
\end{table*}
\section{Experiments}
\label{sec:experiments}
The following section describes the experimental setting used to validate our occlusion inference approach. We set the number of latent classes to $K = 100$ based on computational time and tractability of the baselines. 

\noindent\textbf{Data Processing.}
To train and test our model, we use data from the GL unsignalized intersection in the INTERACTION dataset~\cite{interaction}, which contains highly interactive real-world traffic trajectories from a BeV. We subsample vehicles that serve as the ego vehicle from the available tracks. We then build OGMs around these vehicles, focusing on mapping the region in front of the drivers. We construct the ground truth OGM $\mathcal{M}_{ego}^{gt}$ from the BeV data and the occluded OGM $\mathcal{M}_{ego}^{obs}$ using ray tracing. The ego OGMs are of dimension $70 \times 60$ with a grid cell resolution of \SI{1}{\meter}. Partially observed vehicles are mapped as fully observed, and only completely occluded vehicles are marked as occluded in $\mathcal{M}_{ego}^{obs}$. 
For all visible drivers, their position, orientation, velocity, and acceleration are extracted to form the trajectory vector $s_{h}^{1:T}$ over \SI{1}{\second} of past data sampled at \SI{10}{\hertz}. We filter out all visible drivers not observed for a contiguous \SI{1}{\second} from the dataset. An OGM $\mathcal{M}_{h}$ is constructed for the region ahead of each visible driver $h$ of dimension $20 \times 30$ with a \SI{1}{\meter} resolution. Since we consider a relatively small area in front of the driver sensor, we assume $\mathcal{M}_{h}$ to not have any occlusions.

\noindent\textbf{Baselines.}
We baseline our proposed method against the original PaS approach, referred to as \emph{k-means PaS}~\cite{people_as_sensors}. This algorithm first clusters past driver trajectory data using k-means. Then given a cluster, the occupancy probability of each cell in $\hat{\mathcal{M}}_{h}$ for an observed driver $h$ is computed. Since the original work did not consider multiple observed drivers, we incorporate this driver sensor model into our sensor fusion pipeline to perform multi-agent occlusion inference.

K-means PaS~\cite{people_as_sensors} does not model the multimodality of the inferred environment given a trajectory. Thus, we also consider an extension that uses a Gaussian mixture model (GMM) learned with the expectation-maximization~(EM)~\cite{moon1996expectation} algorithm to cluster the trajectories, termed \emph{GMM~PaS}. The comparison of our approach to GMM~PaS highlights the performance improvements gained from the more expressive CVAE model and learning the OGM end-to-end directly from the trajectory input, rather than first computing the clusters, and then separately inferring the OGM.%

\noindent\textbf{Metrics.}
We consider three core metrics in our evaluation: accuracy, mean squared error (MSE), and image similarity~(IS)~\cite{image_similarity}. To compute accuracy, we threshold the occupancy probability above $0.6$ to be occupied, below $0.4$ to be free, and those in between to be unknown. The accuracy and MSE metrics both focus on precision. 
Since we cannot expect precise recovery of the environment solely from observed driving behavior, we also consider the IS metric as in~\cite{people_as_sensors}. 
IS captures the relative structure of the OGMs using Manhattan distance rather than a one-to-one comparison of individual cells.
Due to occupancy class imbalance, we break down each metric per occupied and free class, and overall. 

To evaluate the ability of GMM PaS and our proposed approach to accurately represent the multimodality of the data, we also consider a multimodal variant of each metric. We take the best metric across the three most likely modes of the GMM and CVAE for the driver sensor model evaluation. 
We do the same for the full occlusion inference pipeline according to the likelihoods $p(\hat{\mathcal{M}}_{ego})$ described in \cref{sec:methods}. We refer to these metrics as ``Top 3''.
\section{Results}
\label{sec:results}
\noindent\textbf{Driver Sensor Model.}
We compare our proposed CVAE driver sensor model to k-means and GMM PaS baselines in \cref{tab:sensor_model}. In almost all metrics, the CVAE outperforms the baselines. 
Despite  the  difficulty  of  the  task  in  terms  of precision, our model was able to achieve high top 3 accuracy.
It outperformed the GMM PaS baseline across all top 3 metrics for the driver sensor model, reflecting better modeling of the multimodality in the distribution. 
The only metric on which the CVAE underperforms is MSE for the occupied class. We hypothesize this is in part due to the CVAE better capturing multimodality in its latent distribution, resulting in the most likely mode not necessarily being the most accurate mode (multiple modes may have similar likelihoods when multimodality is present).

\begin{figure}[t!]
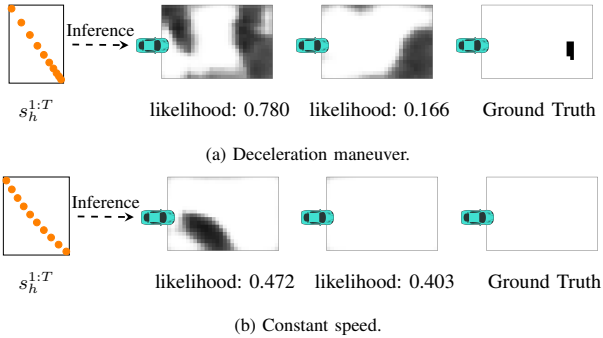

     \centering
     \begin{subfigure}[h!]{\columnwidth}
         \centering
         \scalebox{0.85}{\input{Figures/Qualitative/DriverSensorModel/deceleration}}
         \caption{Deceleration maneuver.}
         \label{fig:dec}
     \end{subfigure}
     \hfill
     \begin{subfigure}[h!]{\columnwidth}
         \centering
         \scalebox{0.85}{\input{Figures/Qualitative/DriverSensorModel/constantspeed}}
         \caption{Constant speed.}
         \label{fig:constant_speed}
     \end{subfigure}
        \caption{Qualitative CVAE driver sensor model results for a deceleration maneuver and a constant speed trajectory. For $\SI{1}{\second}$ input trajectories $s_{h}^{1:T}$, the two most likely inferred OGMs $\hat{\mathcal{M}}_{h}$ are shown, alongside the ground truth OGM $\mathcal{M}_{h}$. The OGMs depict free (white), unknown (gray), and occupied (black) space ahead of the driver. The model results align with our intuition: an observed deceleration likely indicates occupied space ahead, whereas a constant speed profile may indicate either traffic or free space ahead.\vspace{-20pt}}
        \label{fig:results_driver_model}
\end{figure}

\cref{fig:results_driver_model} shows example outputs from the CVAE model for an input deceleration maneuver (\cref{fig:dec}) and a constant speed trajectory (\cref{fig:constant_speed}). Intuitively, the driver in \cref{fig:dec} is likely slowing down due to another road agent ahead or near them. 
The CVAE model reflects this, inferring occupied space ahead in its two most likely latent modes.
For the driver traveling at a constant speed (\cref{fig:constant_speed}), we do not have enough information to conclude if the space ahead of them is occupied (moving with traffic) or free (an open road). Our model correctly captures the multimodality of the scenario inferring either free or occupied space with similar likelihood. These qualitative examples show that the CVAE is able to infer the environment ahead of a driver given their trajectory and correctly capture the associated multimodality.%
\vspace{-10pt}
\begin{figure*}[t!]
     \centering
     \begin{subfigure}[h!]{0.95\columnwidth}
         \centering
         \vspace{5pt}
         \includegraphics[width=0.45\linewidth,trim={50cm 19.5cm 27cm 17cm},clip]{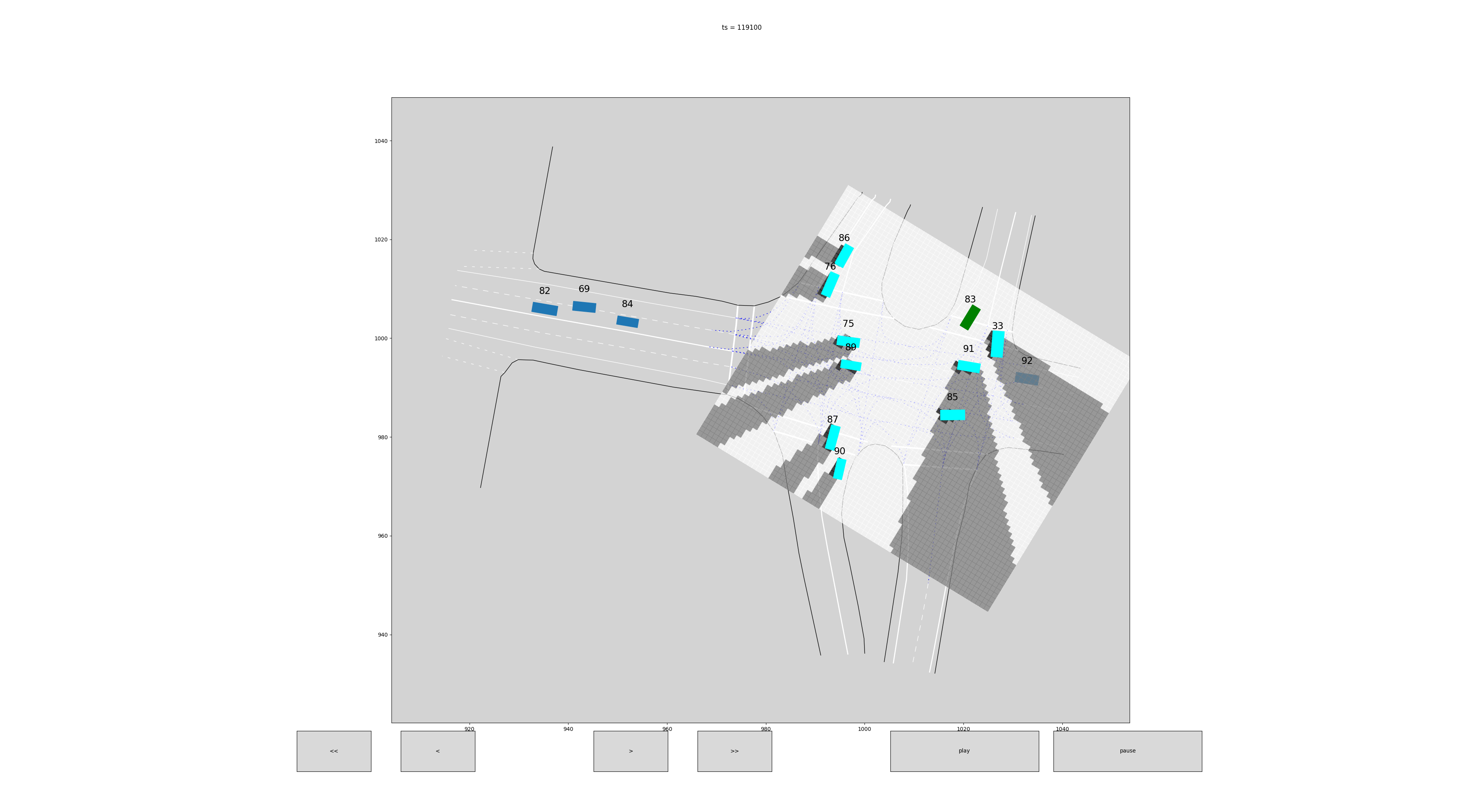}
         \includegraphics[width=0.45\linewidth,trim={50cm 19.5cm 27cm 17cm},clip]{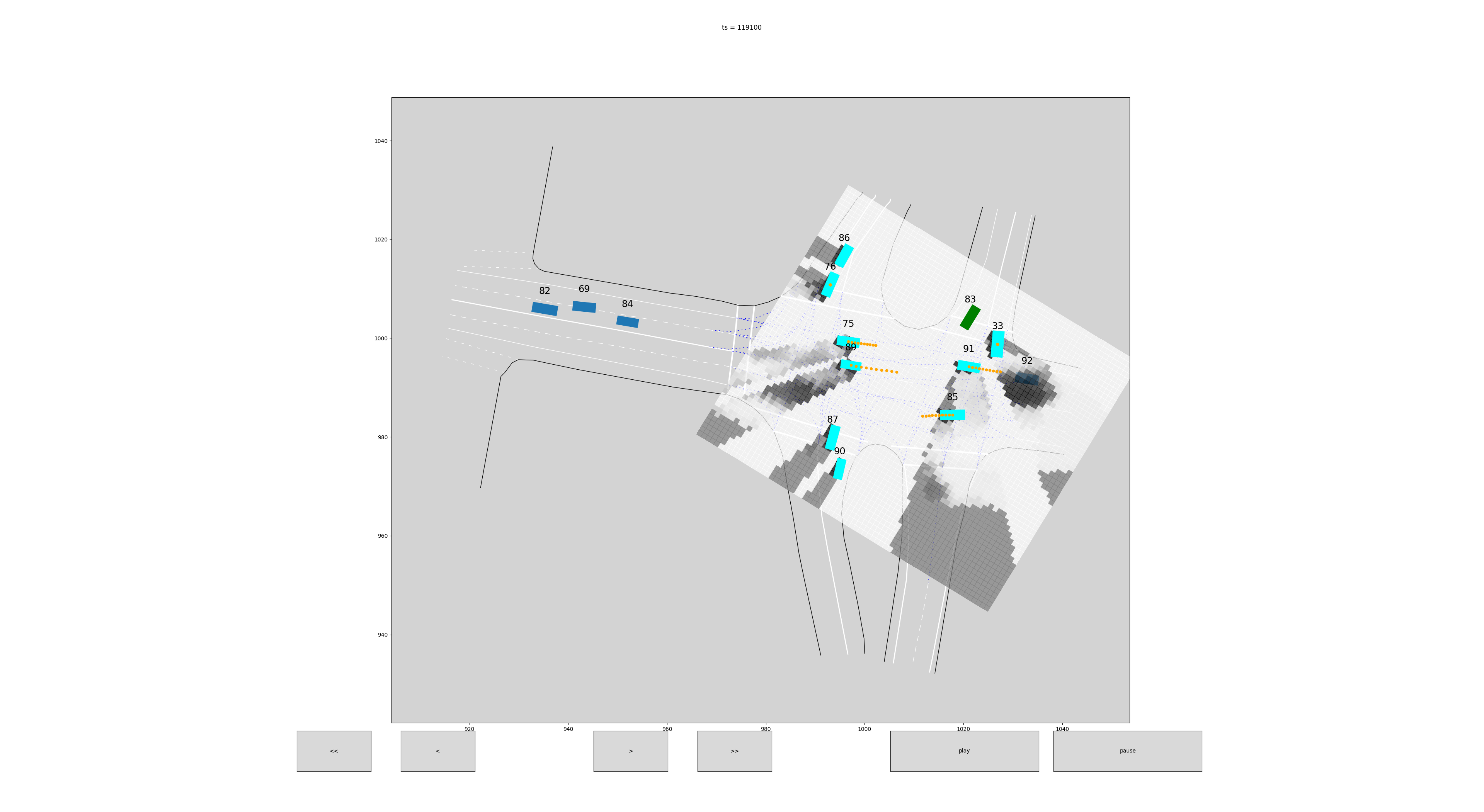}
         \vspace{-3pt}
         \caption{Stopped driver.}
         \label{fig:stopped}
         \vspace{2pt}
     \end{subfigure}
     \begin{subfigure}[h!]{0.95\columnwidth}
         \centering
         \vspace{5pt}
         \includegraphics[width=0.45\linewidth,trim={50cm 19.5cm 27cm 17cm},clip]{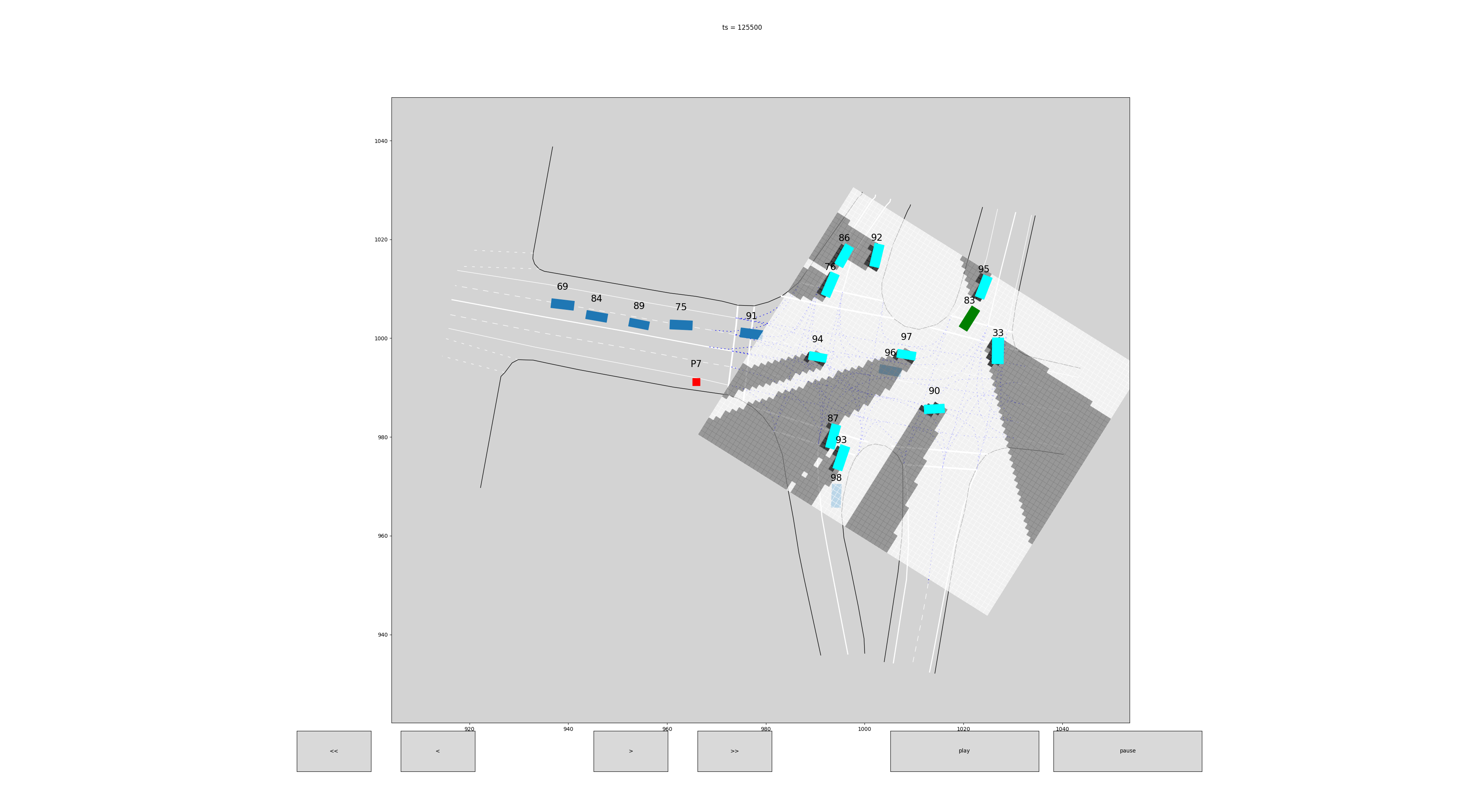}
         \includegraphics[width=0.45\linewidth,trim={50cm 19.5cm 27cm 17cm},clip]{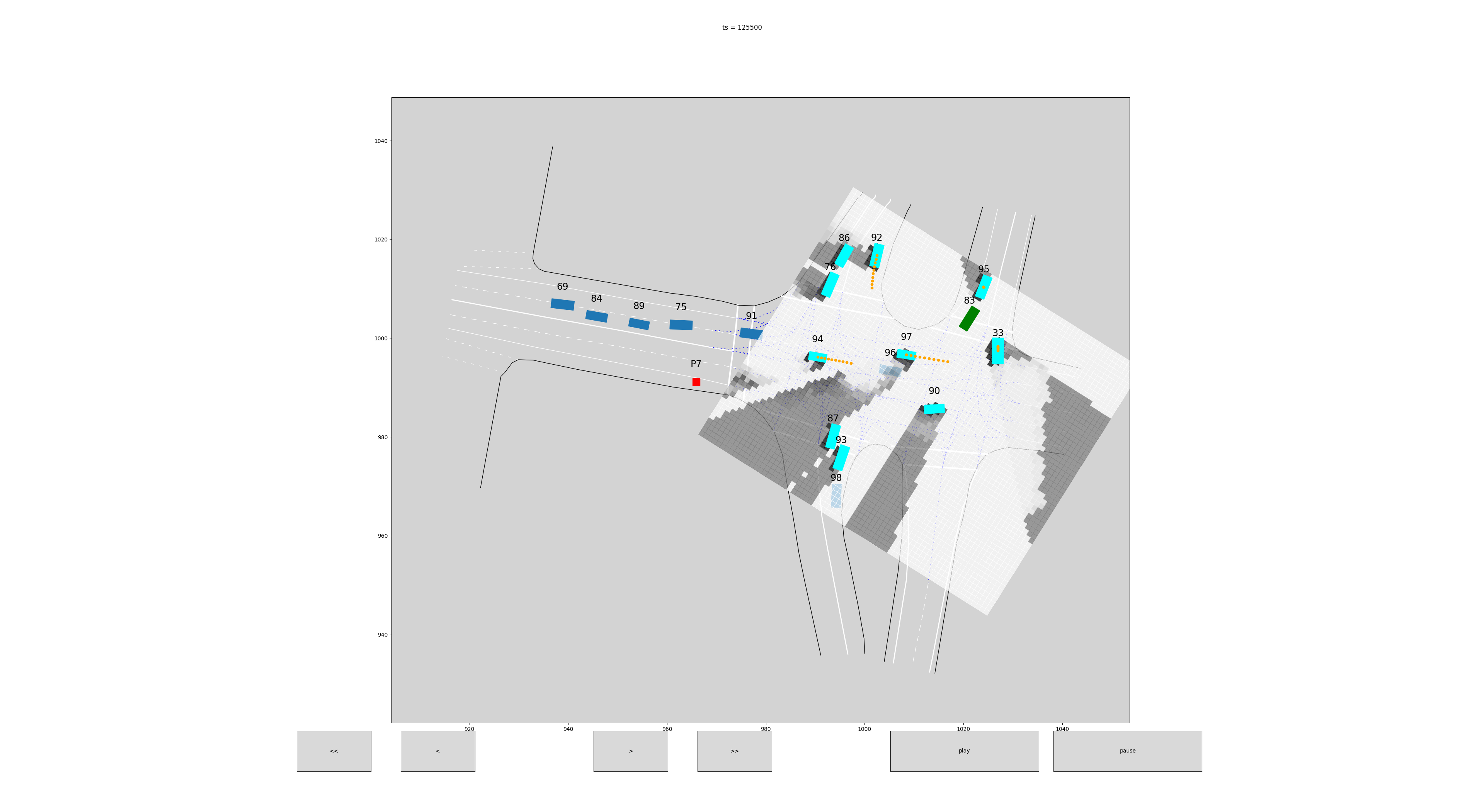}
         \vspace{-3pt}
         \caption{Accelerating driver.}
         \label{fig:moving}
         \vspace{2pt}
     \end{subfigure}
     \begin{subfigure}[h!]{0.95\columnwidth}
         \centering
         \includegraphics[width=0.45\linewidth,trim={51cm 20cm 29cm 17cm},clip]{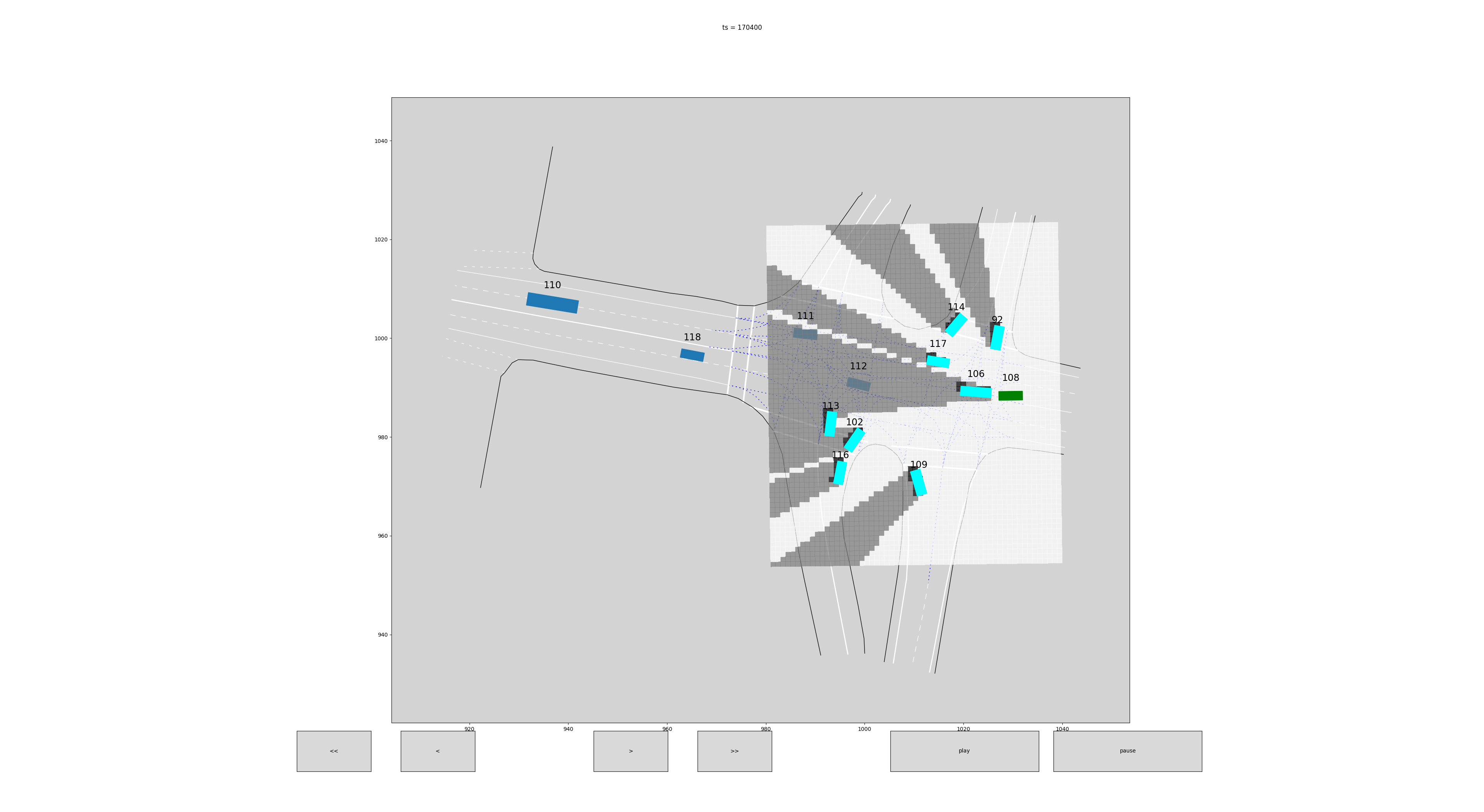}
         \includegraphics[width=0.45\linewidth,trim={51cm 20cm 29cm 17cm},clip]{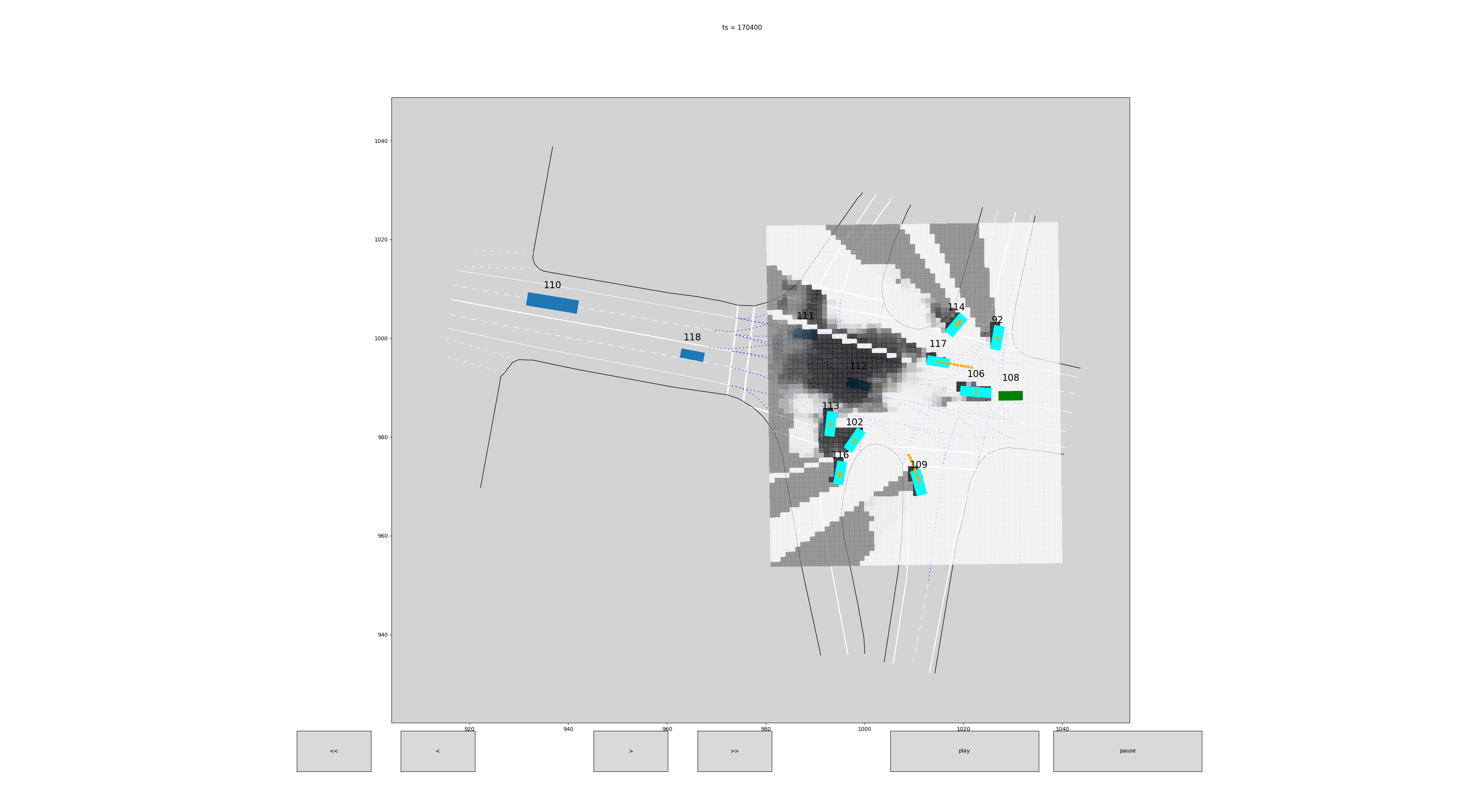}
         \vspace{-3pt}
         \caption{Multiple stopped drivers.}
         \label{fig:multiple_stopped}
     \end{subfigure}
     \begin{subfigure}[h!]{0.95\columnwidth}
         \centering
         \includegraphics[width=0.45\linewidth,trim={51cm 20cm 29cm 17cm},clip]{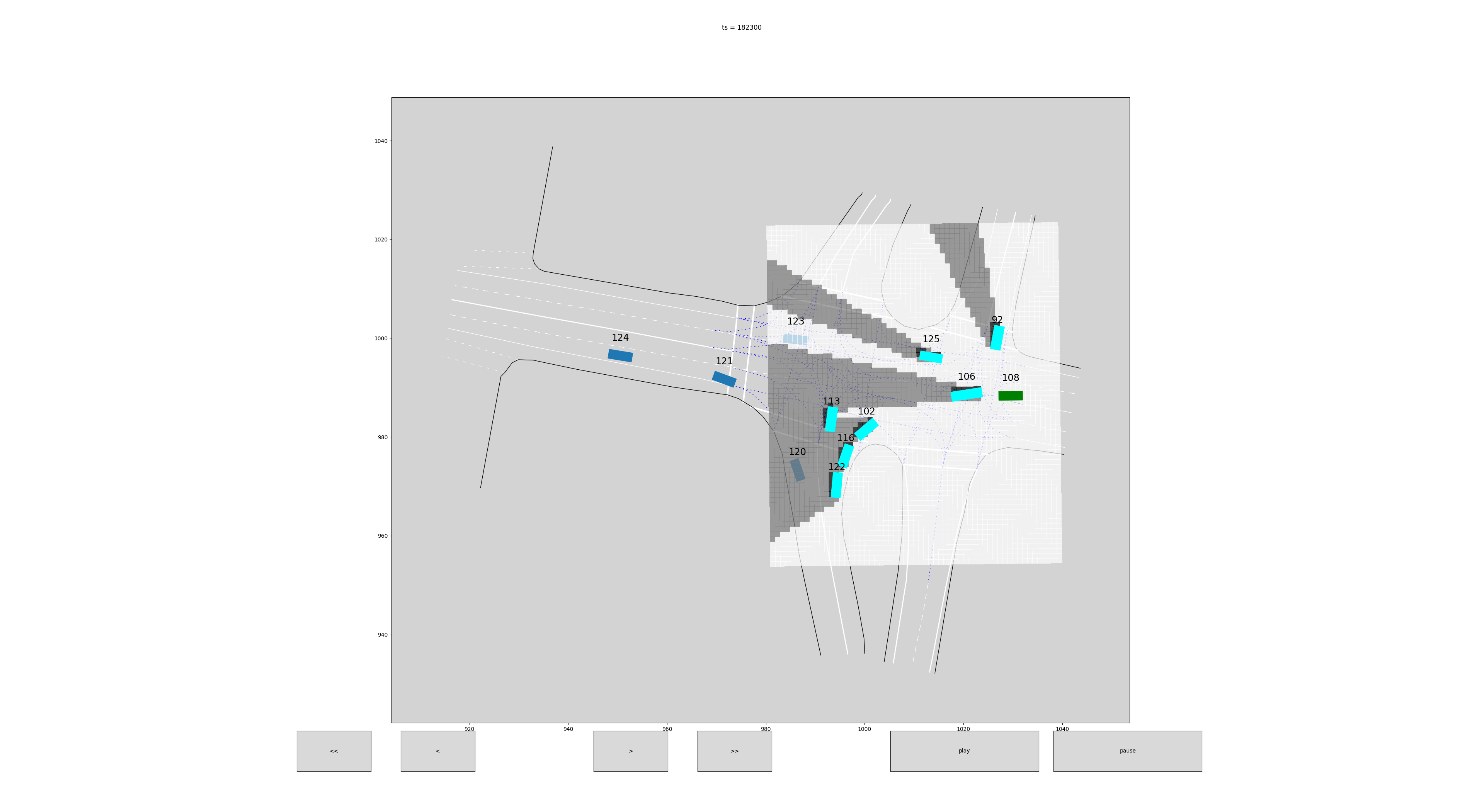}
         \includegraphics[width=0.45\linewidth,trim={51cm 20cm 29cm 17cm},clip]{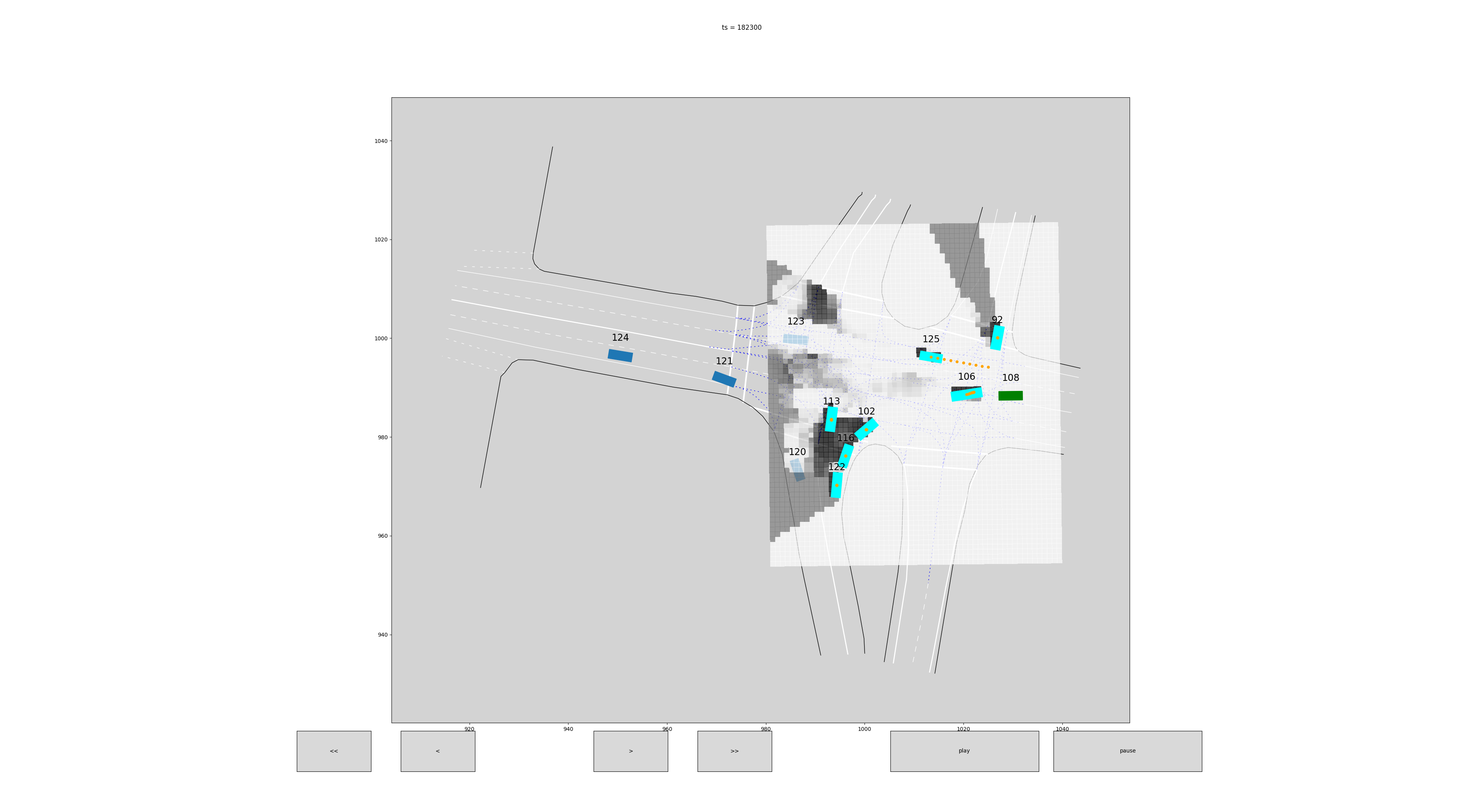}
         \vspace{-3pt}
         \caption{Multiple stopped drivers with one accelerating driver.}
         \label{fig:multiple_accel}
     \end{subfigure}
        \caption{Qualitative results for the full pipeline using our proposed CVAE driver sensor model. The scenarios in \cref{fig:stopped} and \cref{fig:moving} depict the ego vehicle (green) waiting to make a right turn. In \cref{fig:multiple_stopped,fig:multiple_accel}, the ego vehicle is waiting to make a left turn. The observed drivers are shown in cyan and their trajectories for the past \SI{1}{\second} ($s_{h}^{1:T}$) in orange. The vanilla OGM $\mathcal{M}_{ego}^{obs}$ (left) and the fused OGM $\hat{\mathcal{M}}_{ego}$ (right) depict free (white), occluded (gray), and occupied (black) space around the ego. The occupancy patterns inferred by the algorithm match our intuition based on the observed behaviors.
        % \vspace{-2pt}
        }
        \label{fig:results_full_pipeline}
\end{figure*}
\begin{figure*}[t!]
     \centering
     \begin{subfigure}[h!]{0.49\columnwidth}
         \centering
         \includegraphics[width=0.85\columnwidth,trim={49cm 26cm 28cm 18cm},clip]{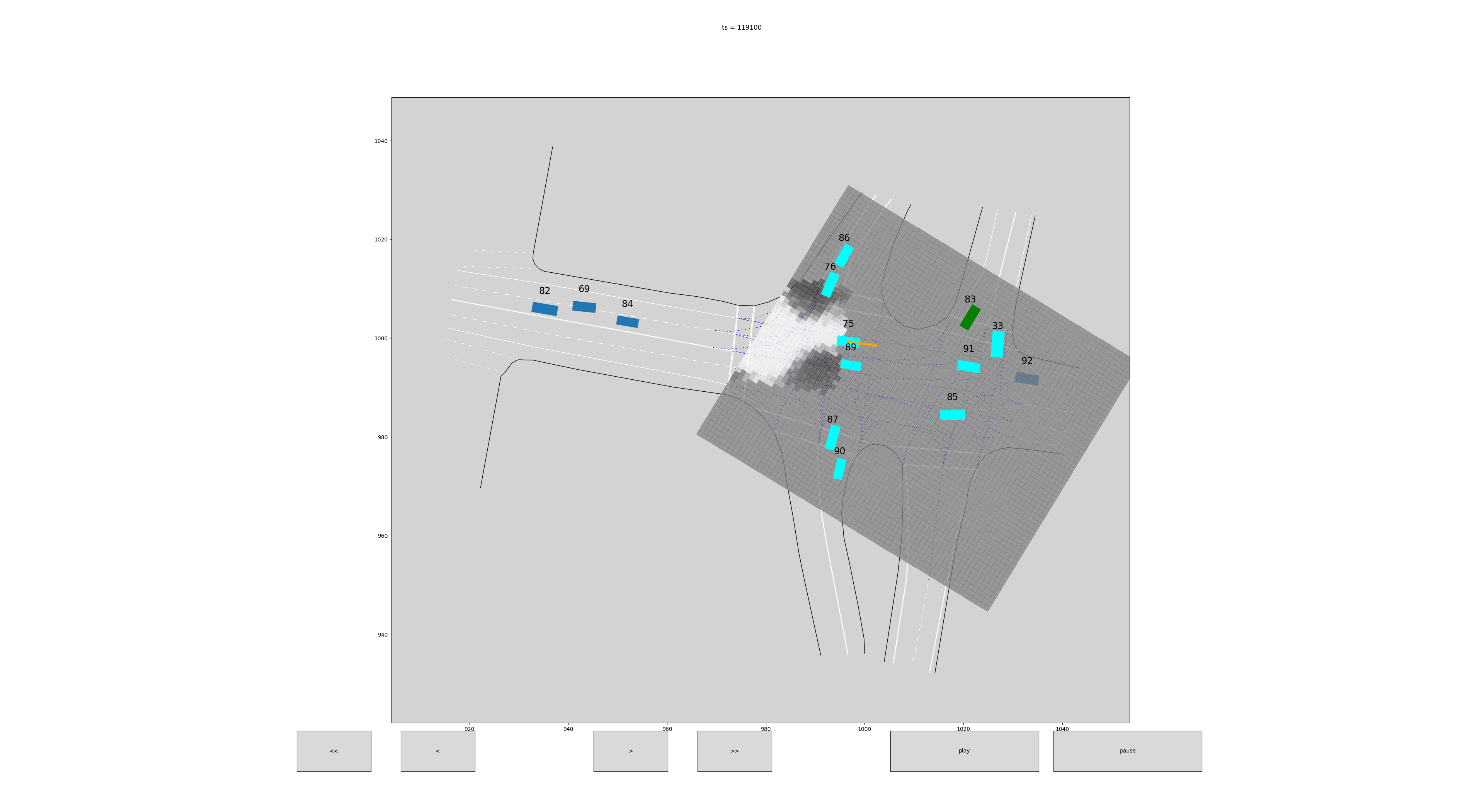}\vspace{-3pt}
         \caption{Driver 75: mode 1.
         (likelihood: 0.423).
         }
         \label{fig:vehicle_75_mode_1_only}
         \vspace{5pt}
     \end{subfigure}
     \begin{subfigure}[h!]{0.49\columnwidth}
         \centering
         \includegraphics[width=0.85\columnwidth,trim={49cm 26cm 28cm 18cm},clip]{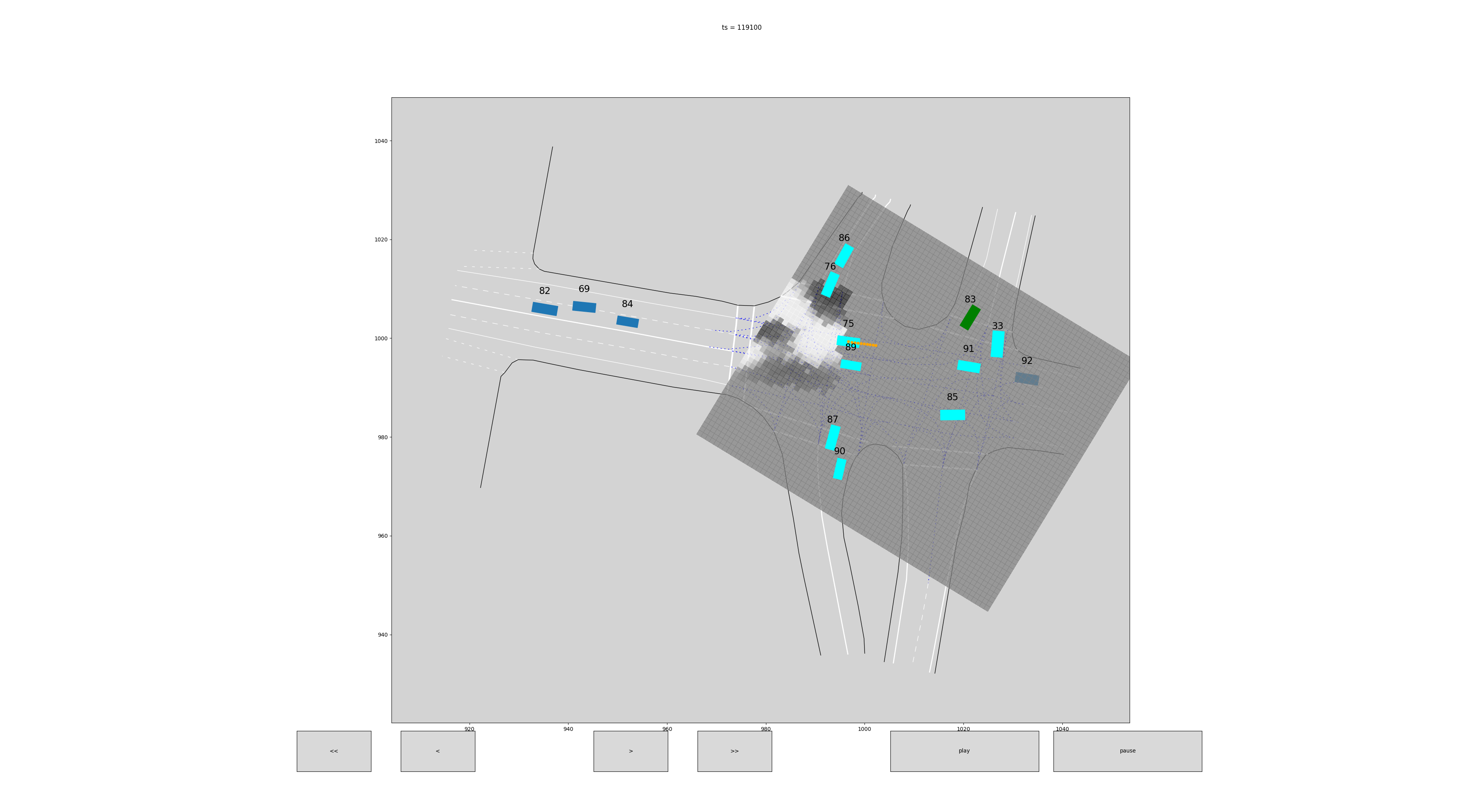}\vspace{-3pt}
         \caption{Driver 75: mode 2.
         (likelihood: 0.410).
        }
         \label{fig:vehicle_75_mode_2_only}
         \vspace{5pt}
     \end{subfigure}
     \begin{subfigure}[h!]{0.49\columnwidth}
         \centering
         \includegraphics[width=0.85\columnwidth,trim={49cm 26cm 28cm 18cm},clip]{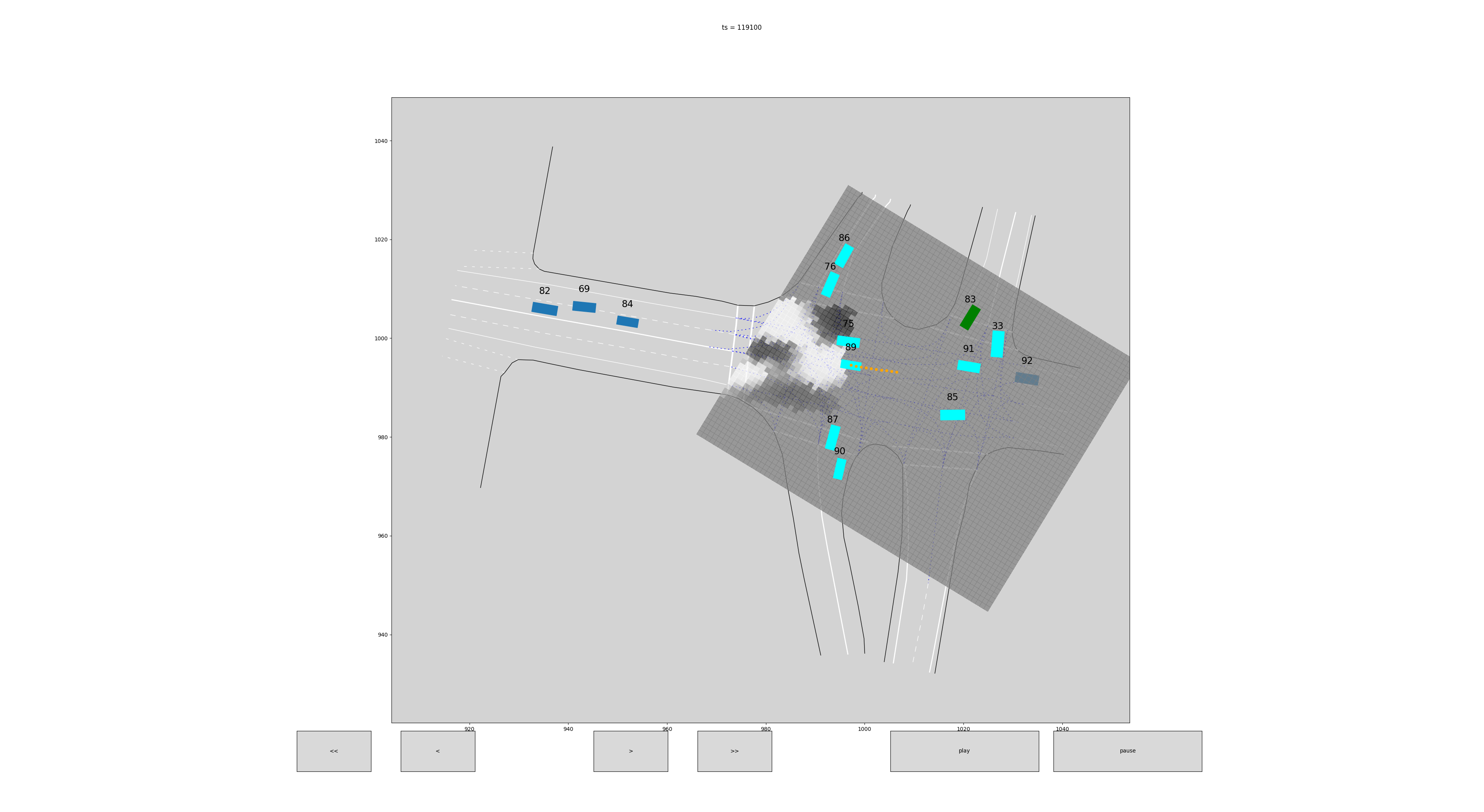}\vspace{-3pt}
         \caption{Driver 89: mode 1.
         (likelihood: 0.652).
        }
         \label{fig:vehicle_89_mode_1_only}
         \vspace{5pt}
     \end{subfigure}
     \begin{subfigure}[h!]{0.49\columnwidth}
         \centering
         \includegraphics[width=0.85\columnwidth,trim={49cm 26cm 28cm 18cm},clip]{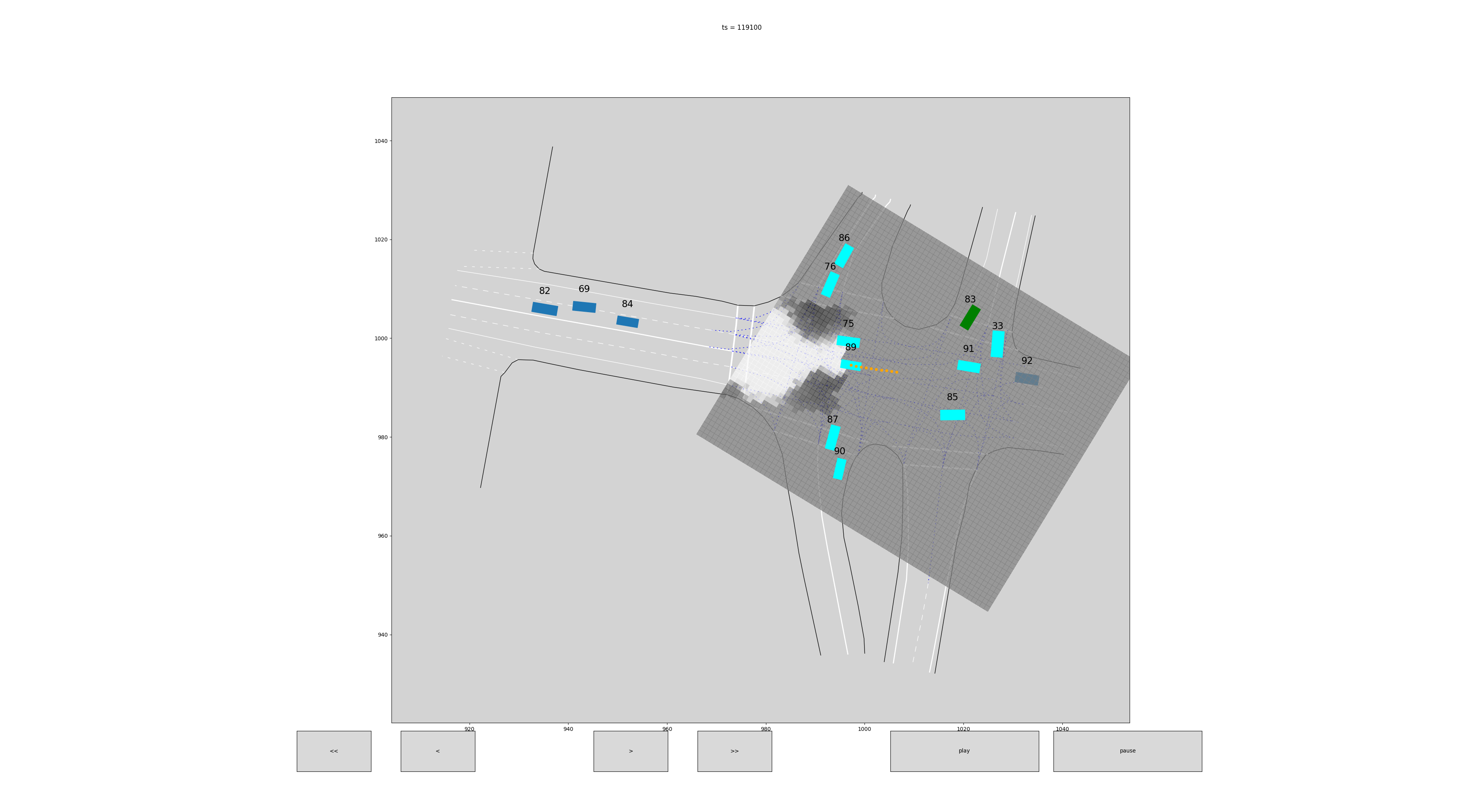}\vspace{-3pt}
         \caption{Driver 89: mode 2.
         (likelihood: 0.114).
        }
         \label{fig:vehicle_89_mode_2_only}
         \vspace{5pt}
     \end{subfigure}
     \begin{subfigure}[h!]{0.49\columnwidth}
         \centering
         \includegraphics[width=0.85\columnwidth,trim={49cm 26cm 28cm 18cm},clip]{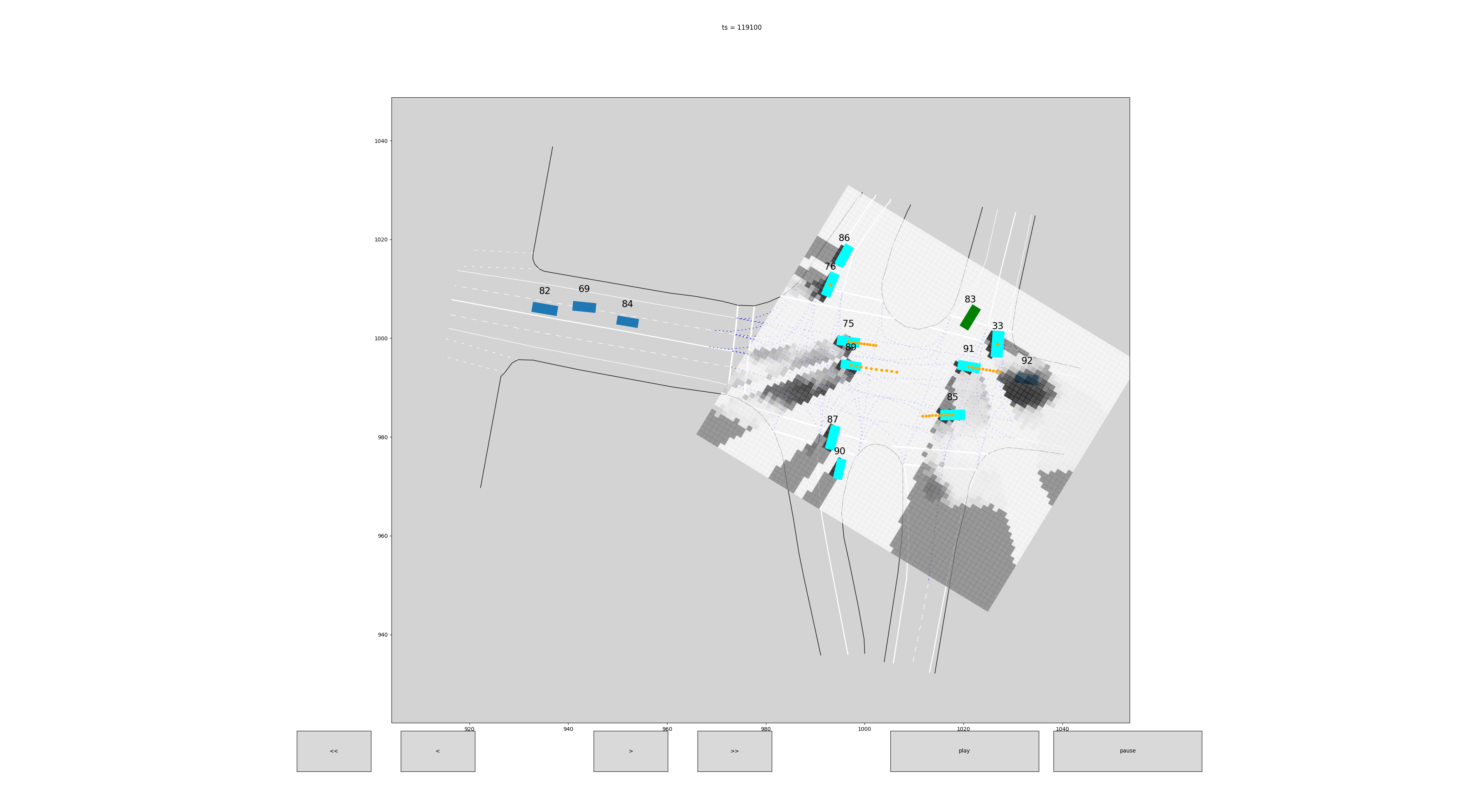}\vspace{-3pt}
         \caption{Driver 75: mode 1, Driver 89: mode 1.}
         \label{fig:vehicle_75_mode_1}
     \end{subfigure}
     \begin{subfigure}[h!]{0.49\columnwidth}
         \centering
         \includegraphics[width=0.85\columnwidth,trim={49cm 26cm 28cm 18cm},clip]{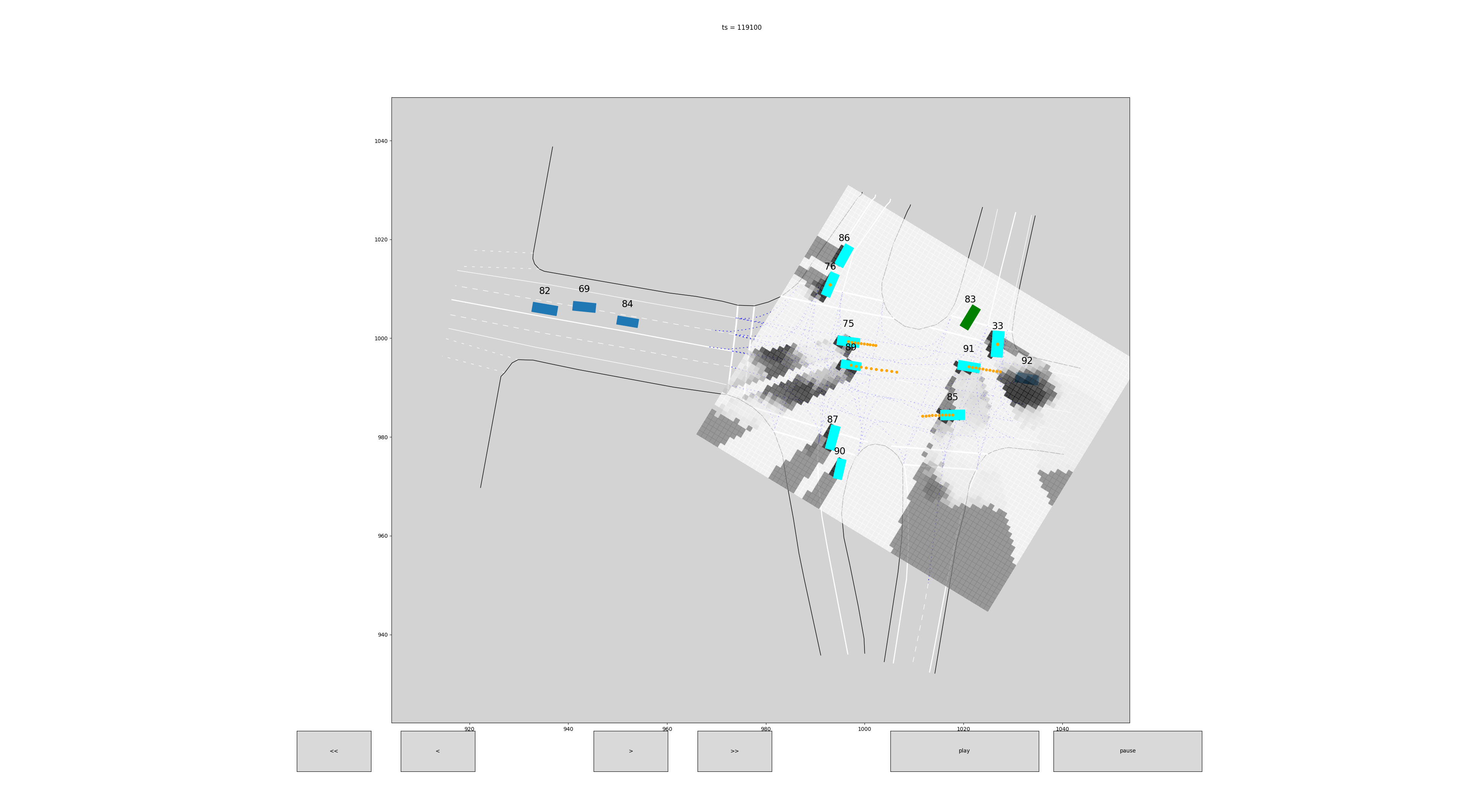}\vspace{-3pt}
         \caption{Driver 75: mode 2, Driver 89: mode 1.}
         \label{fig:vehicle_75_mode_2}
     \end{subfigure}
     \begin{subfigure}[h!]{0.49\columnwidth}
         \centering
         \includegraphics[width=0.85\columnwidth,trim={49cm 26cm 28cm 18cm},clip]{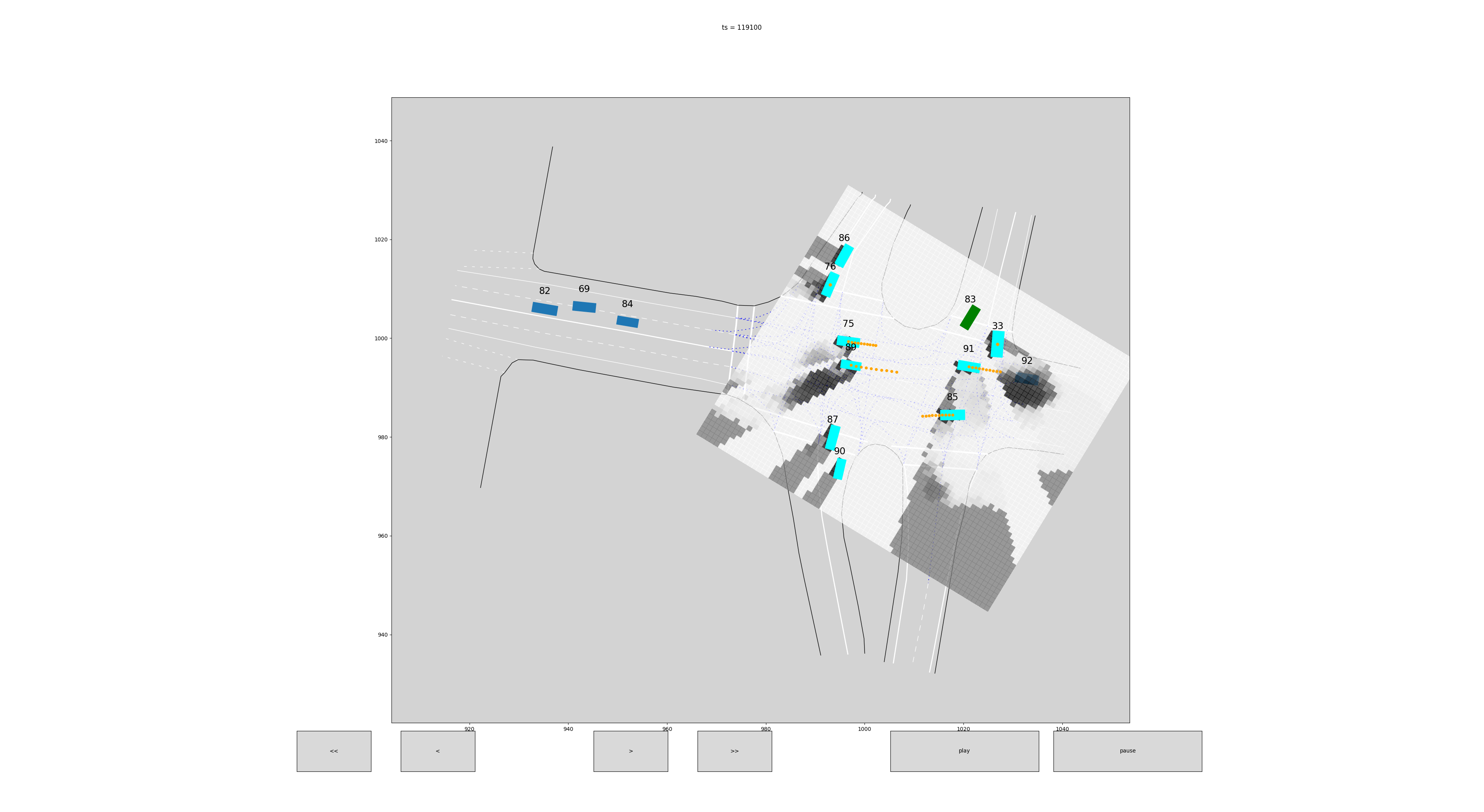}\vspace{-3pt}
         \caption{Driver 75: mode 1, Driver 89: mode 2.}
         \label{fig:vehicle_89_mode_2}
     \end{subfigure}
     \begin{subfigure}[h!]{0.49\columnwidth}
         \centering
         \includegraphics[width=0.85\columnwidth,trim={49cm 26cm 28cm 18cm},clip]{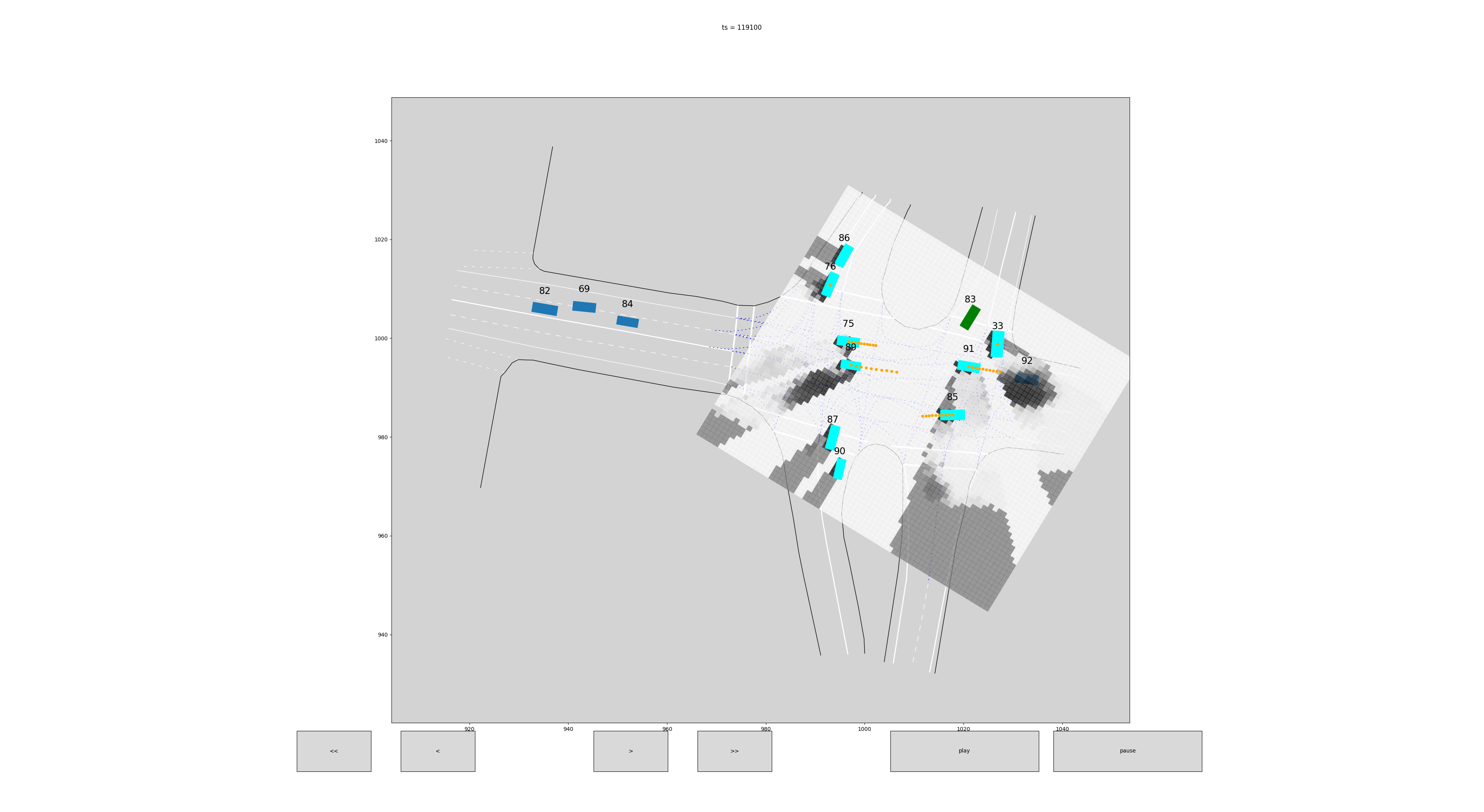}\vspace{-3pt}
         \caption{Driver 75: mode 2, Driver 89: mode 2.}
         \label{fig:vehicle_75_89_mode_2}
     \end{subfigure}
        \caption{Multimodal results for the full occlusion inference pipeline
        for the scenario in \cref{fig:stopped,fig:moving}. The ego vehicle is shown in green, the observed drivers in cyan, their trajectories for the past \SI{1}{\second} ($s_{h}^{1:T}$) in orange, and the occluded vehicle in blue. The OGMs depict free (white), occluded (gray), and occupied (black) space. 
        Observed drivers 75 and 89 are moving at a constant speed. 
        \cref{fig:vehicle_75_mode_1_only,fig:vehicle_75_mode_2_only,fig:vehicle_89_mode_1_only,fig:vehicle_89_mode_2_only} show the top two most likely modes for the inferred OGMs $\hat{\mathcal{M}}_{75}$ and $\hat{\mathcal{M}}_{89}$ within a fully occluded ego vehicle OGM for easier visualization.
        Both drivers' top two most likely modes depict either free or occupied space ahead (following our intuition from \cref{fig:constant_speed}). 
        \cref{fig:vehicle_75_mode_1,fig:vehicle_75_mode_2,fig:vehicle_89_mode_2,fig:vehicle_75_89_mode_2} depict the top two most likely modes for drivers 75 and 89 in the fused OGM $\hat{\mathcal{M}}_{ego}$. 
        The inferred OGMs output by our algorithm accurately capture the multimodal possibilities for the occluded space ahead of the observed drivers.
        % \vspace{-15pt}
        }
        \label{fig:results_full_pipeline_multimodal}
\end{figure*}

\noindent\textbf{Multi-Agent Occlusion Inference.}
We now consider the full occlusion inference pipeline, incorporating the inferences from multiple drivers into $\mathcal{M}_{ego}^{obs}$ using the evidential sensor fusion mechanism to form $\hat{\mathcal{M}}_{ego}$. 
Our proposed approach with the CVAE driver sensor model and evidential sensor fusion runs on average at 
\SI{56}{\hertz}, which is real-time capable given that typical perception systems operate at \SI{10}{\hertz}~\cite{guizilini2019dynamic}. 
We compute the metrics across only the occluded cells in $\mathcal{M}_{ego}^{obs}$ and ignore the cells in $\hat{\mathcal{M}}_{ego}$ that are thresholded to $0.5$. 
\cref{tab:sensor_model} shows we outperform baselines on all metrics except accuracy and MSE for the occupied class. Since we only incorporate the inferred occupancy in regions of occlusion for the ego vehicle, this biases the subset of grid cells that are updated (e.g., to the side of the vehicle as for driver 75 in \cref{fig:stopped}). For these regions, the baselines lean more toward the occupied class than the CVAE. Nevertheless, the IS metric conveys that the structure of the occupancy in the occluded region is best captured by the CVAE model.%

\cref{fig:results_full_pipeline} illustrates scenarios with different observed driver behaviors. In \cref{fig:stopped,fig:moving}, the ego vehicle (green) is waiting for a gap in traffic in order to turn right at the unsignalized intersection. Observed driver 33 is blocking the ego vehicle's view of oncoming traffic while waiting to turn left. In \cref{fig:stopped}, driver 33 is stopped due to occluded driver~92 passing by.
Our approach correctly infers occupied space in the region near driver 92. In \cref{fig:moving}, driver 33 begins to turn left due to a gap in traffic. 
Our method successfully infers free space ahead of driver 33, indicating to the ego vehicle that it may be safe to proceed with its right turn. 
In \cref{fig:multiple_stopped,fig:multiple_accel}, the ego vehicle is waiting to turn left at the unsignalized intersection behind driver 106, who is blocking the view of oncoming traffic. There are multiple drivers visible to the ego vehicle. In \cref{fig:multiple_stopped}, 
observed drivers 106, 113, 102, and 116 are all stopped due to occluded driver 112 passing by. The CVAE driver sensor model outputs occupied space ahead for each of these observed drivers. Our multi-agent occlusion inference algorithm thus infers that the occluded area is occupied with high probability after fusing the measurements from all driver sensors, correctly identifying occupied space around occluded driver 112. In \cref{fig:multiple_accel}, driver 106 begins their left turn. Our method successfully infers free space ahead of driver 106, indicating to the ego vehicle that it may be safe to proceed with its own maneuver (e.g., a U-turn or a left turn). In all cases, our approach is able to provide additional insight about the scene to the ego vehicle as compared to the vanilla OGM $\mathcal{M}_{ego}^{obs}$ with no occlusion inference.%

However, not all observed trajectories provide sufficient information to precisely infer the occupancy ahead of the driver (e.g., constant speed driving), making the ability to capture the multimodality of the different options for these trajectories important. 
In \cref{fig:results_full_pipeline_multimodal}, we consider two observed constant speed drivers: 75 and 89 from \cref{fig:stopped,fig:moving} and their top two most likely inferred modes. We expect the learned distribution to capture the possibility that there is free or occupied space ahead of these drivers, corresponding to driving in an empty lane or in free-flow traffic. 
\cref{fig:vehicle_75_mode_1_only,fig:vehicle_75_mode_2_only,fig:vehicle_89_mode_1_only,fig:vehicle_89_mode_2_only} show that the top two decoded modes are the same for both drivers. One mode depicts free space ahead with occupied space to the sides (i.e., an open lane), and the other mode depicts occupied space ahead (i.e., traffic ahead). The CVAE outputs almost equal likelihoods for these two modes for driver 75. For driver 89, occupied space ahead is more likely, potentially, due to the dataset statistics in this location of the intersection favoring traffic ahead of the driver.%

In \cref{fig:vehicle_75_mode_1,fig:vehicle_75_mode_2,fig:vehicle_89_mode_2,fig:vehicle_75_89_mode_2}, we analyze how the multimodality for these constant speed drivers manifests itself in the full occlusion inference pipeline. We consider all possible combinations of the two most likely OGMs for drivers 75 and 89. For the remaining observed drivers, we use their most likely OGMs. In \cref{fig:vehicle_75_mode_2}, both drivers 75 and 89 produce OGMs that depict traffic ahead, resulting in high inferred occupancy levels ahead of these vehicles in the fused OGM $\hat{\mathcal{M}}_{ego}$. In \cref{fig:vehicle_89_mode_2}, both drivers produce OGMs that depict free space ahead, thus their combination in the fused OGM $\hat{\mathcal{M}}_{ego}$ yields the lowest occupancy probability ahead. In practice, to ensure tractability, a set number of most likely OGMs $\hat{\mathcal{M}}_{ego}$ can be considered by a path planner. We hypothesize that incorporating the multimodal options into a planner would result in more proactive and informed behavior by the ego vehicle. High multimodality in the distribution is also an indicator of insufficient information provided by the observed driver behavior to confidently infer the occluded occupancy.%
\section{Conclusion}
\label{sec:conclusion}
We proposed a real-time capable, multi-agent framework for occlusion inference that uses observed human behaviors to inform occluded regions of an ego vehicle's map. Our approach captures human-like intuition for reasoning about occlusions under uncertainty. Experiments show that accurately modeling the multimodality of the data distribution improves performance in this setting. 
There are several interesting avenues for future work. To focus the driver sensor model on only the highly interactive behaviors, ideas from anomaly detection~\cite{chandola2009anomaly} may be used. Moreover, incorporating HD map information into the driver sensor model can allow semantic reasoning (e.g., based on navigable space or traffic light state) to inform occlusion inference.%

\newpage

% \section*{ACKNOWLEDGMENT}
% {\small We thank Ransalu Senanayake for the enlightening discussions. We thank Spencer M. Richards for his invaluable feedback. This project was made possible with funding from the Ford-Stanford Alliance and a gift from Mercedes-Benz Research \& Development North America.}

{\small
\printbibliography
}

% \clearpage
\appendix

\subsection{Experimental Details}
\label{sec:appendix_experiments}
All experiments were run on an Intel Core i7-6700HQ processor and an NVIDIA 8GB GeForce RTX 1070 GPU. 

\subsubsection{Data Split}
\label{subsec:appendix_data}
To process the data, ego vehicle IDs were sampled from the GL intersection in the INTERACTION dataset~\cite{interaction}. For each of the 60 available scenes, a maximum of 100 ego vehicles were chosen. The train, validation, and test sets were randomly split based on the ego vehicle IDs accounting for $85\%$, $5\%$ and $10\%$ of the total number of ego vehicles, respectively. Due to computational constraints, the training set for the driver sensor model was further reduced to have 70,001 contiguous driver sensor trajectories or 2,602,332 time steps of data. The validation set used to select the driver sensor model and the sensor fusion scheme contained 4,858 contiguous driver sensor trajectories or 180,244 time steps and $289$ ego vehicles. The results presented in this paper were reported on the test set, which consists of 9,884 contiguous driver sensor trajectories or 365,201 time steps and $578$ ego vehicles.

\subsubsection{Example OGMs}
\begin{figure}[h!]
     \centering
      \begin{subfigure}[h!]{0.235\textwidth}
         \centering
         \includegraphics[width=\textwidth,trim={4cm 1cm 3.5cm 1cm},clip]{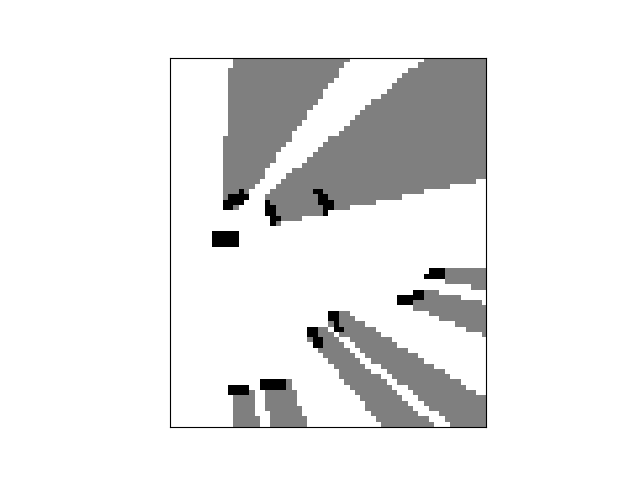}
         \caption{$\mathcal{M}_{ego}^{obs}$}
     \end{subfigure}
     \begin{subfigure}[h!]{0.235\textwidth}
         \centering
         \includegraphics[width=\textwidth,trim={4cm 1cm 3.5cm 1cm},clip]{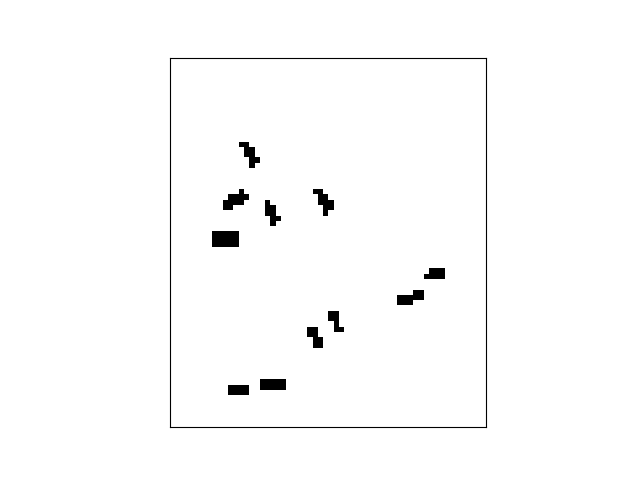}
         \caption{$\mathcal{M}_{ego}^{gt}$}
     \end{subfigure}
        \caption{\small Example ego vehicle OGMs. They depict free (white), occluded (gray), and occupied (black) space around the ego vehicle, which is located in the middle left of each OGM. $\mathcal{M}_{ego}^{obs}$ contains occluded regions, whereas $\mathcal{M}_{ego}^{gt}$ is omniscient.
        }
        \label{fig:example_OGMs}
\end{figure}

We show exemplary ego OGMs in \cref{fig:example_OGMs}. These OGMs are from the same time step as \cref{fig:stopped}. Examples of inferred driver sensor and ground truth OGMs can be seen in \cref{fig:results_driver_model}. We intentionally chose the dimensions of the driver sensor OGMs ($20 \times 30$) to be smaller than the dimensions of the ego OGMs ($70 \times 60$) to reflect that the driver sensors are likely reacting to the local environment closest to them.

\subsubsection{CVAE Driver Sensor Model Architecture and Training}
We set the number of latent classes in the CVAE to $K~=~100$ based on computational time and tractability for the considered baselines. We standardize the trajectory data $s_{h}^{1:T}$ to have zero mean and unit standard deviation. The prior encoder in the model consists of an LSTM~\cite{lstm} with a hidden dimension of $5$ to process the \SI{1}{\second} of trajectory input data $s_{h}^{1:T}$. A linear layer then coverts the output into a $K$-dimensional vector that goes into a softmax function, producing the prior distribution. The posterior encoder extracts features from the ground truth input $\mathcal{M}_{h}$ using a VQ-VAE backbone~\cite{vqvae} with a hidden dimension of $4$. These features are flattened and concatenated with the LSTM output from the trajectory data, and then passed into a linear layer and a softmax function, producing the posterior distribution. The decoder passes the latent encoding $z$ through two linear layers with ReLU activation functions and a transposed VQ-VAE backbone, outputting the OGM $\hat{\mathcal{M}}_{h}$.

To compute the approximate mutual information in the loss (third term in \cref{eq:loss}), we use the prior ${p_{\theta}(z \!\mid\! s_{h}^{1:T})}$ as a proxy for the posterior distribution and acquire $p_{\theta}(z)$ by summing over the batch following~\cite{salzmann2020trajectron}. To avoid latent space collapse, we clamped the KL divergence term in the loss in \cref{eq:loss} at 0.2. Additionally, we anneal the $\beta$ hyperparameter in the loss according to a sigmoid schedule as recommended by~\citet{bengioKL}. In our hyperparameter search, we found a maximum $\beta$ of $1$ with a crossover point at 10,000 iterations to work well. The $\beta$ hyperparameter increases from a value of $0$ to $1$ over 1,000 iterations. We set the $\alpha$ hyperparameter to $1.5$. We trained the network with a batch size of $256$ for $30$ epochs using the Adam optimizer~\cite{adam} with a starting learning rate of $0.001$.

\subsubsection{Multi-Sensor Fusion Experimental Details}
\label{subsec:params_sensor_fusion}
To perform sensor fusion, a correspondence between the grid cells of an inferred driver sensor OGM $\hat{\mathcal{M}}_{h}$ and the ego vehicle's OGM $\mathcal{M}_{ego}^{obs}$ needs to be found. For each occluded grid cell ($\mathcal{M}_{ego}^{obs}$ = 0.5), the corresponding cell in $\hat{\mathcal{M}}_{h}$ is identified to transfer the occupancy information. We first transform both grids to a global frame. We use a k-d tree~\cite{bentley1975multidimensional} of the grid cells in $\hat{\mathcal{M}}_{h}$ in the global frame to find the nearest neighbor match for each occluded grid cell in $\mathcal{M}_{ego}^{obs}$. Cells that do not have a match have a default distance of infinity. We further set a distance tolerance of \SI{1}{\meter} to ensure only reasonable nearest neighbor cells are found.
If no nearest neighbor cell is identified from $\hat{\mathcal{M}}_{h}$, the occluded grid cell probability remains unchanged in $\mathcal{M}_{ego}^{obs}$ as 0.5. This process is repeated for each observed driver $h$, producing a collection of estimated OGMs $\hat{\mathcal{M}}_{h}$ in the ego vehicle's frame of reference. The collection is then fused into a single updated grid $\hat{\mathcal{M}}_{ego}$ according to the evidential sensor fusion procedure described in \cref{sec:methods}. 

To compute the multimodality metrics for the full occlusion inference pipeline in \cref{tab:sensor_model}, we had to determine the three most likely modes for each time step according to the likelihood of $\hat{\mathcal{M}}_{ego}$: $p(\hat{\mathcal{M}}_{ego})=\prod_{h} p_{\theta}(z \mid s_{h}^{1:T})$ for all visible drivers~$h$. To avoid computing intractably many products for all the modes for each sensor driver, we use breadth-first search to find the three most likely modes for the multimodality metrics.

\subsubsection{Image Similarity Metric}
\label{subsec:appendix_metrics}
The IS metric $\psi$ is computed as follows~\cite{image_similarity}:
\begin{equation}
\begin{aligned}
    \psi(\mathcal{M}_{1},\mathcal{M}_{2}) &= \sum_{a \in \left\{0,1\right\}}d(\mathcal{M}_{1},\mathcal{M}_{2},a) + d(\mathcal{M}_{2},\mathcal{M}_{1},a) \\
	d(\mathcal{M}_{1},\mathcal{M}_{2},a) &= \frac{\sum_{\mathcal{M}_{1,c_{1}} = a} \min \left\{ \Vert c_{1} - c_{2} \Vert_{1} : \mathcal{M}_{2,c_{2}} = a \right\}}{\#_{a}(\mathcal{M}_{1})}
\end{aligned}
\end{equation}
where $\mathcal{M}_{1}$ and $\mathcal{M}_{2}$ are OGMs, $c_{1}$ and $c_{2}$ are the 2D spatial coordinates for each cell in the OGMs $\mathcal{M}_{1}$ and $\mathcal{M}_{2}$, respectively, and $\mathcal{M}_{1,c_{1}}$ is the occupancy class at grid cell $c_{1}$ in OGM $\mathcal{M}_{1}$. Here, $\Vert \cdot \Vert_{1}$ denotes the Manhattan distance between the spatial coordinates, and $\#_{a}(\mathcal{M}_{1})$ is the number of cells in $\mathcal{M}_{1}$ with occupancy class $a$. 

We consider only the grid cells that have an occupancy probability thresholded to occupied (i.e., for probability $\geq 0.6$, $\mathcal{M}_{c} = 1$) or free (i.e., for probability $\leq 0.4$, $\mathcal{M}_{c} = 0$). Unknown cells (i.e., for $0.4 <$~probability~$< 0.6$, $\mathcal{M}_{c} = 0.5$) are ignored in the computation. 

\subsubsection{Details on the K-means PaS and GMM PaS Baselines}
The first step in both the k-means PaS and GMM PaS driver sensor model baselines is to cluster the trajectories $s_{h}^{1:T}$ in the training data. We have $K = 100$ clusters for each model. Then given a cluster $z \in \{1, \hdots, K\}$, the occupancy probability of each cell in $\hat{\mathcal{M}}_{h}$ for an observed driver $h$ is computed. The grid cells are assumed to be independent of each other. For each grid cell~$c$ and cluster $z$, we can compute the occupancy probability in the driver sensor OGM using Bayes' Rule, i.e.,
\begin{equation}
    \begin{aligned}
        \hat{\mathcal{M}}_{h,c} = p(a_{c} = 1 \mid z) = \frac{p(z \mid a_{c} = 1)p(a_{c} = 1)}{p(z)}   
    \end{aligned}
\end{equation}
where $a_{c} \in \{0,1\}$ is the occupancy class. We assume $p(a_{c})$ is uniform over the occupied and free classes (i.e., $p(a_{c})~=~0.5$). From the law of total probability, we have
\begin{equation}
    \begin{aligned}
        \hat{\mathcal{M}}_{h,c} &= p(a_{c} = 1 \mid z)\\ &= \frac{p(z \mid a_{c} = 1)p(a_{c} = 1)}{\sum_{a_{c} \in \{0,1\}}p(z \mid a_{c})p(a_{c})}\\ &= \frac{p(z \mid a_{c} = 1)}{p(z \mid a_{c} = 1) + p(z \mid a_{c} = 0)}.
    \end{aligned}
\end{equation}
The term $p(z \mid a_{c})$ is computed by dividing the number of occurrences of $z$ and $a_{c}$ together in the training set by the number of occurrences of $a_{c}$ according to the definition of conditional probability.

\subsection{Results}
\label{sec:appendix_results}
\subsubsection{Decoded Latent Classes}
\begin{figure}[h!]
     \centering
     \begin{subfigure}[h!]{0.47\textwidth}
         \centering
         \includegraphics[width=\textwidth,trim={3cm 5cm 2cm 5cm},clip]{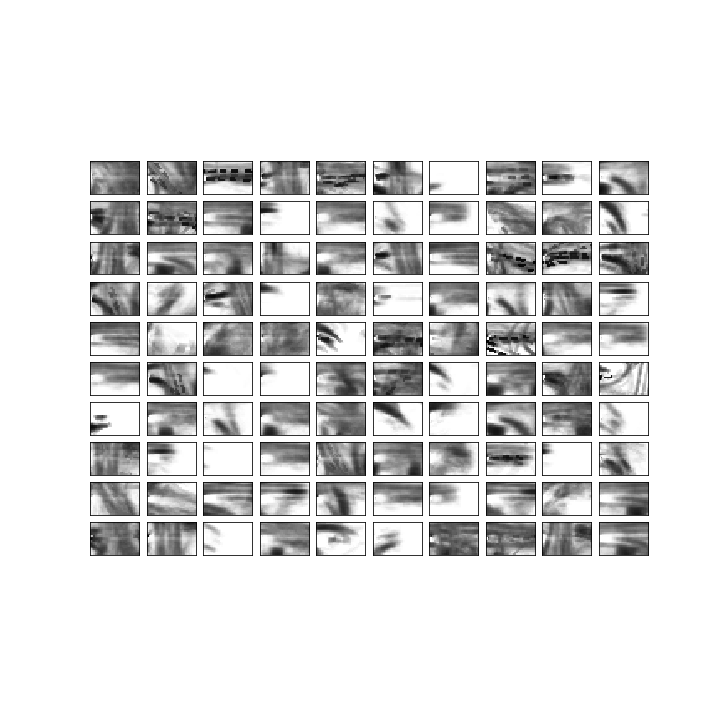}
         \caption{K-means PaS}
     \end{subfigure}
     \begin{subfigure}[h!]{0.47\textwidth}
         \centering
         \includegraphics[width=\textwidth,trim={3cm 5cm 2cm 5cm},clip]{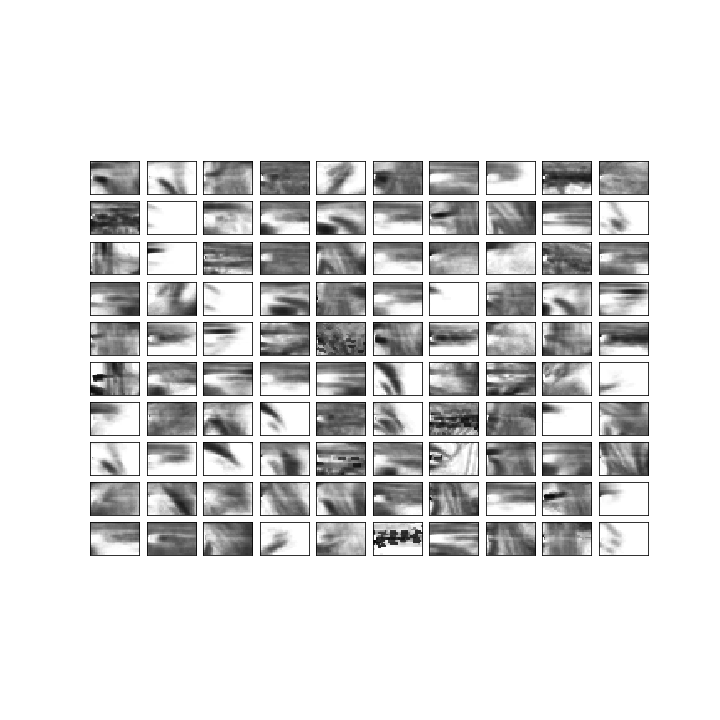}
         \caption{GMM PaS}
     \end{subfigure}
     \begin{subfigure}[h!]{0.47\textwidth}
         \centering
         \includegraphics[width=\textwidth,trim={3cm 5cm 2cm 5cm},clip]{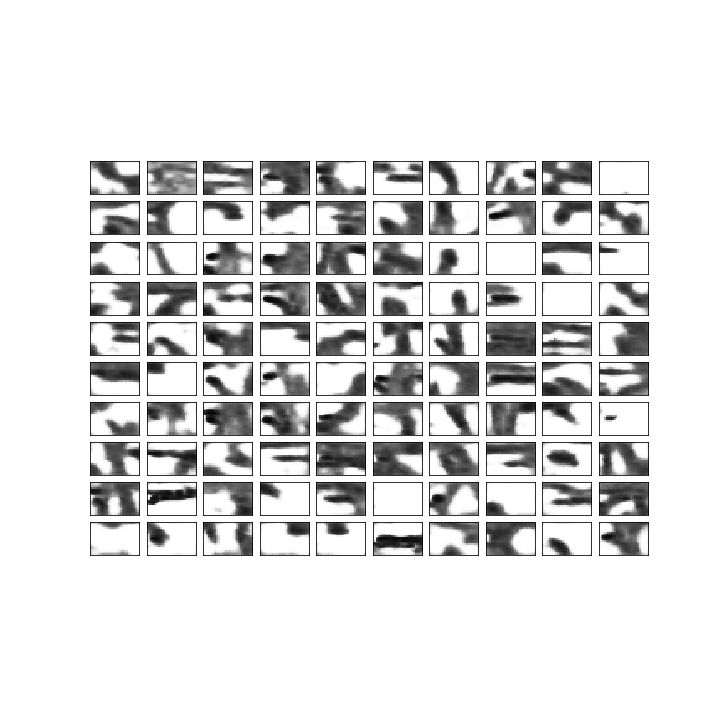}
         \caption{Ours}
     \end{subfigure}
        \caption{\small Decoded latent classes from the considered driver sensor models. Our model produces higher occupancy values in more localized regions than the baselines, which tend to infer mostly lower valued occupied or unknown space.
        }
        \label{fig:decoded_latent_spaces}
\end{figure}

The $K = 100$ decoded latent classes for each of the considered driver sensor models: k-means PaS, GMM PaS, and CVAE (ours), are shown in \cref{fig:decoded_latent_spaces}. The latent spaces of k-means PaS and GMM PaS are quite similar as can be expected due to the structural similarities between both models. However, the CVAE is distinct from the baselines in that the decoded occupancy values appear higher (likely due to the weighted reconstruction loss that favors occupied classes) and more localized. There is a higher number of decoded OGMs for the CVAE that only have localized occupancy regions as opposed to mostly occupied or unknown space as in the k-means PaS and GMM PaS models. Generally, the baselines tend to output more unknown space in the OGMs than the CVAE.

\subsubsection{Additional Multi-Agent Occlusion Inference Baselines}
\label{subsec:appendix_additional_baseline}
\begin{table}[b!]
\begin{center} 
\caption{\footnotesize Ablation study for the multi-agent occlusion inference pipeline. Our evidential fusion method outperforms baseline approaches across most metrics for the full occlusion inference pipeline. Bold denotes the best performing model across a metric. Note that IS values are divided by 100. The maximum standard error per grid cell in MSE and accuracy metrics is 0.0008, and per OGM in the IS metric is 0.0046.} \label{tab:average_baseline}
\footnotesize
\begin{tabular}{l|rrr|rrr}
\hline
& Occ. & Free & Overall & Occ. & Free & Overall \\
\hline\hline

Method & \multicolumn{3}{c|}{Acc. $\uparrow$} &
\multicolumn{3}{c}{Top 3 Acc. $\uparrow$} \\
\hline
Vanilla OGM & 0 & 0 & 0 & N/A & N/A & N/A \\
Ours (Avg.) & \textbf{0.693} & 0.721 & 0.721 & \textbf{0.762} & 0.770 & 0.768 \\
Ours (Evid.) & 0.660 & \textbf{0.722} & \textbf{0.722} & 0.746 & \textbf{0.778} & \textbf{0.774} \\
\hline

& \multicolumn{3}{c|}{MSE $\downarrow$} &
\multicolumn{3}{c}{Top 3 MSE $\downarrow$} \\
\hline
Vanilla OGM & \textbf{0.250} & 0.250 & 0.250 & N/A & N/A & N/A \\
Ours (Avg.) & 0.269 & \textbf{0.166} & \textbf{0.167} & \textbf{0.214} & \textbf{0.135} & \textbf{0.138} \\
Ours (Evid.) & 0.303 & 0.171 & 0.173 & 0.233 & 0.136 & 0.140 \\
\hline

& \multicolumn{3}{c|}{IS $\downarrow$} &
\multicolumn{3}{c}{Top 3 IS $\downarrow$} \\
\hline
Vanilla OGM & \textbf{0.974} & 2.600 & 3.573 & N/A & N/A & N/A \\
Ours (Avg.) & 1.345 & 0.019 & 1.364 & 1.229 & 0.013 & 1.243 \\
Ours (Evid.) & 1.336 & \textbf{0.017} & \textbf{1.353} & \textbf{1.220} & \textbf{0.011} & \textbf{1.232} \\
\hline 
\end{tabular}
\end{center}
\end{table}
We consider two additional baselines for the multi-agent occlusion inference pipeline: a naive averaging strategy in place of the proposed evidential sensor fusion approach and the vanilla OGM $\mathcal{M}_{ego}^{obs}$ which has $0.5$ probability in occluded cells and does not have any occlusion inference. The results of these two baselines and our proposed method with evidential sensor fusion for reference are shown in \cref{tab:average_baseline}. This table is an extension of \cref{tab:sensor_model}.

The vanilla OGM baseline serves as a sanity check, but does not provide meaningful metric results as all the occupancy values are set to $0.5$. This yields an accuracy of $0$ and an MSE of $0.25$. In the IS metric, the constant occupancy value of $0.5$ manifests itself in having the maximum Manhattan distance for the free and occupied cells in the ground truth OGM. However, the IS metric includes distances from both the inferred OGM and the ground truth OGM, which serves as an advantage for the vanilla OGM. The distances from the inferred OGM to the ground truth grid cells are $0$ since 0.5 values are ignored in the computation (see \cref{subsec:appendix_metrics}), thus lowering the IS metric value. The metrics are computed over the occluded regions in $\mathcal{M}_{ego}^{obs}$ and where the k-means PaS baseline does not output unknown space. The vanilla OGM outperforms our proposed methods for the MSE and IS metrics in the occupied class according to \cref{tab:average_baseline}. The MSE metric is biased to $0.5$ values, preferring them over a more confident, but imprecise occupancy output. The IS metric is artificially lower for the vanilla OGM in the occupied class because the ground truth has relatively few occupied cells, and there are no inferred occupied cells in the vanilla OGM, unlike the driver sensor models.

Interestingly, the naive averaging strategy performed comparably to evidential sensor fusion in \cref{tab:average_baseline}. A potential reason for this could be the spatial bias introduced by including the occlusion inference only in regions of occlusion from the ego vehicle. These regions often do not contain the interaction for the driver sensor (e.g., to the side of the driver). Averaging outperformed evidential sensor fusion on the MSE metric. We hypothesize that this might be due to the averaged inferred values being closer to 0.5 than those from the evidential sensor fusion. This leads to improvement in the MSE metric as it is biased towards 0.5 probabilities. In line with this hypothesis, we found in our experiments that the averaging approach resulted in a slightly higher number of inferred cells being thresholded to 0.5 as compared to the evidential sensor fusion technique. In the IS metric, our method with evidential sensor fusion outperformed averaging, showing better captured structure of the inferred environment.

\subsubsection{Individual Inferred OGMs for Observed Drivers}
\begin{figure*}[h!]
     \centering
     \begin{subfigure}[h!]{0.47\textwidth}
         \centering
         \includegraphics[width=\textwidth,trim={46.1cm 13cm 19cm 12cm},clip]{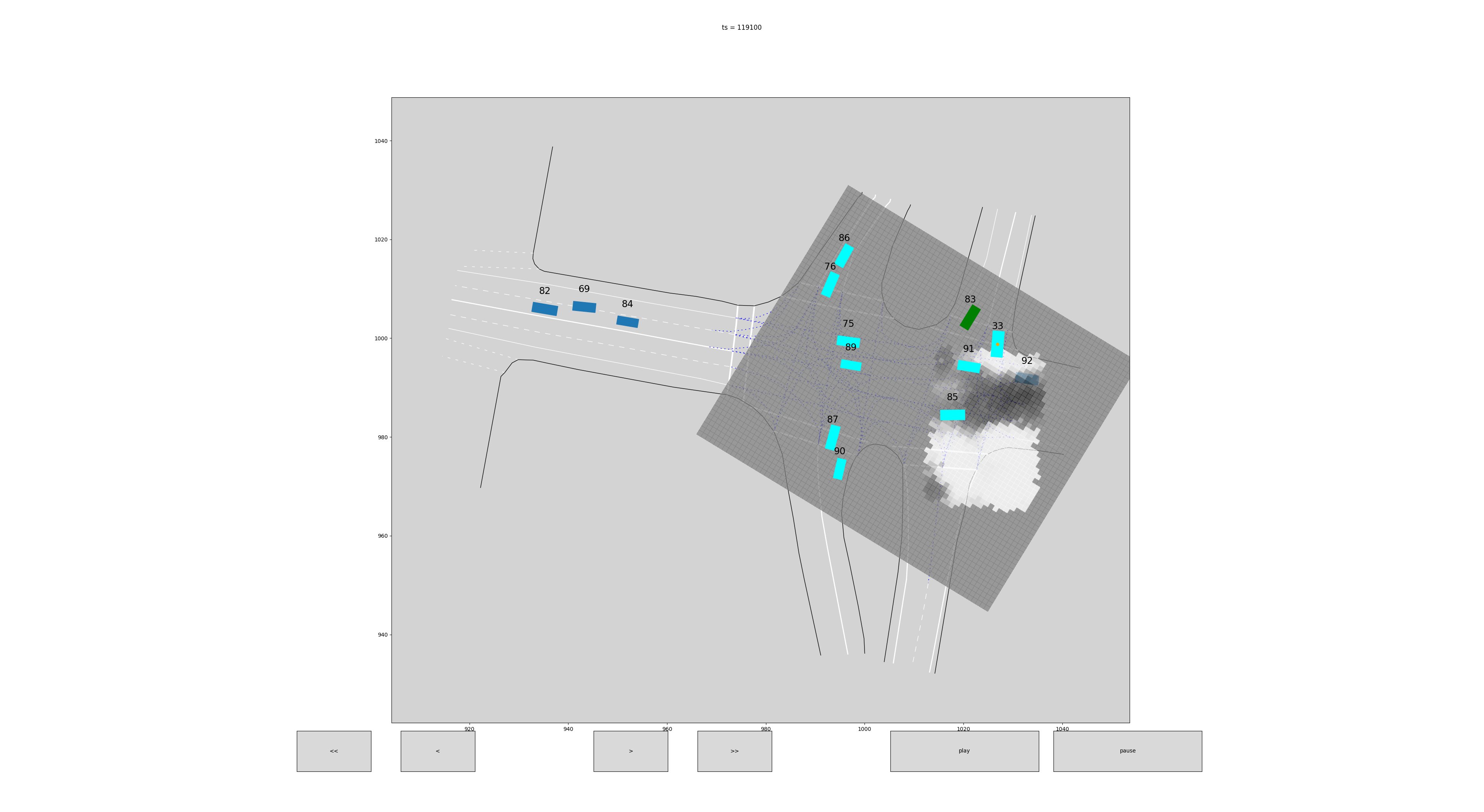}
         \caption{Vehicle 33: stopped.}
         \label{fig:vehicle_33_stopped}
     \end{subfigure}
     \begin{subfigure}[h!]{0.47\textwidth}
         \centering
         \includegraphics[width=\textwidth,trim={46.1cm 13cm 19cm 12cm},clip]{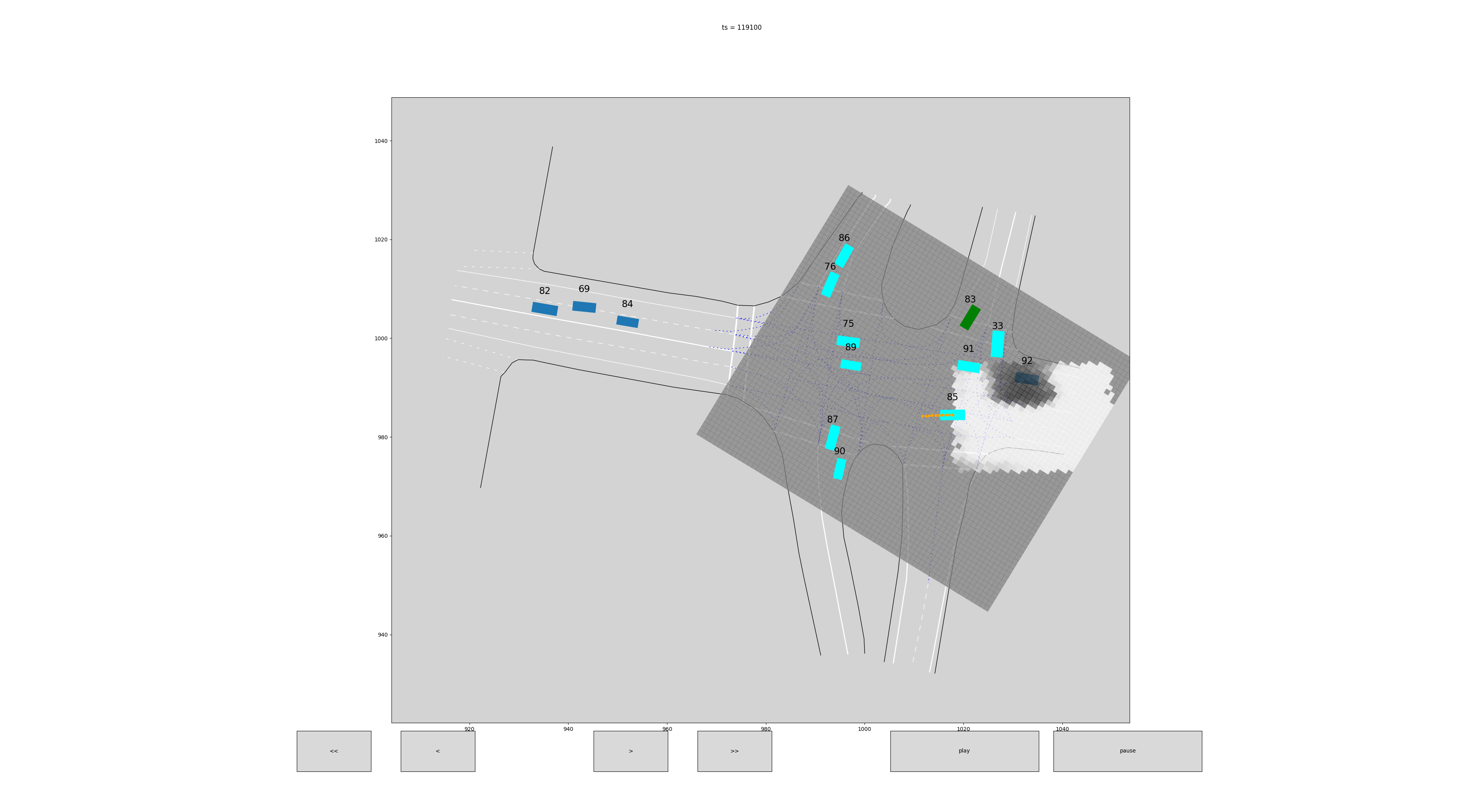}
         \caption{Vehicle 85: constant speed.}
         \label{fig:vehicle_85}
     \end{subfigure}
     \begin{subfigure}[h!]{0.47\textwidth}
         \centering
         \includegraphics[width=\textwidth,trim={46.1cm 13cm 19cm 12cm},clip]{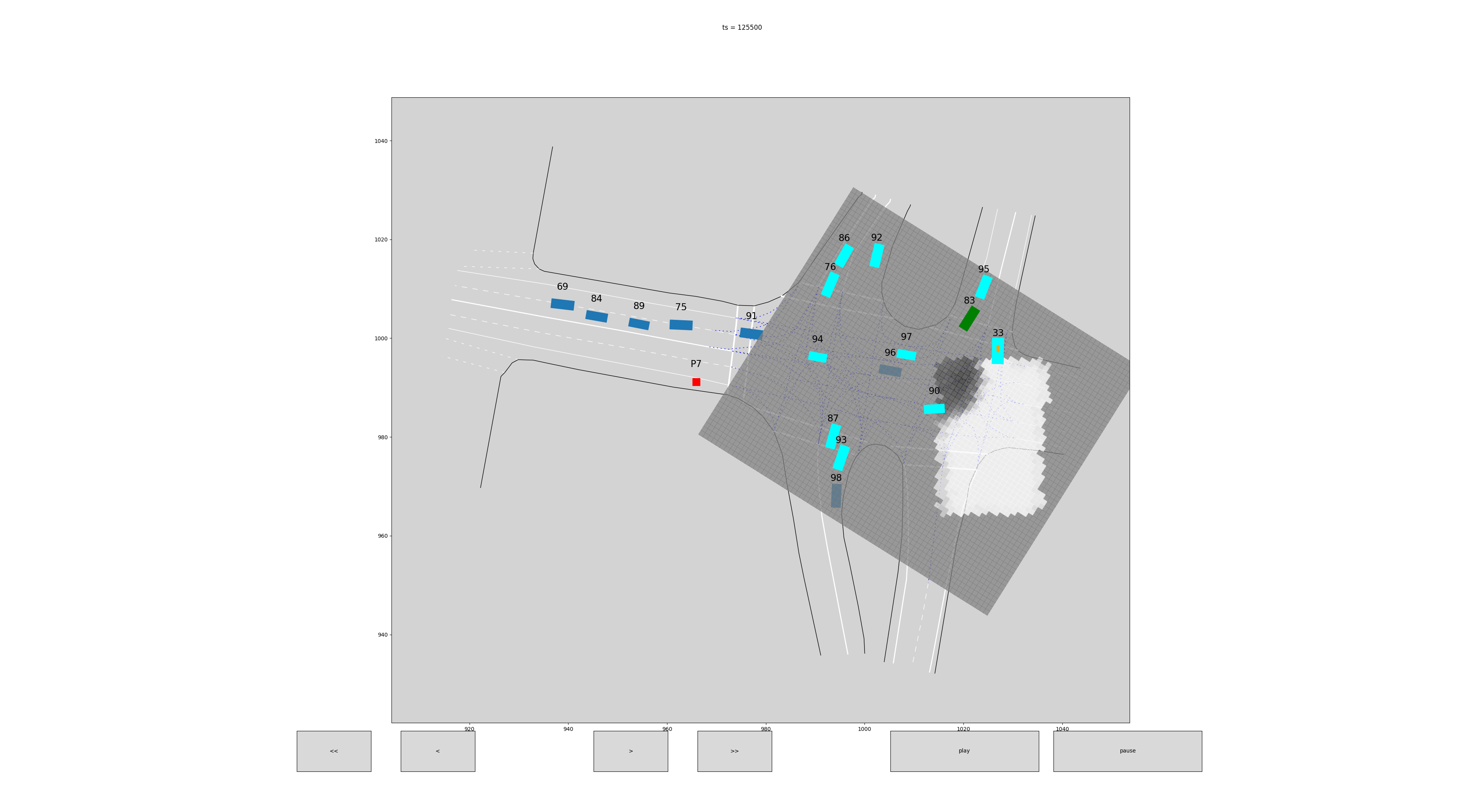}
         \caption{Vehicle 33: accelerating.}
         \label{fig:vehicle_33_acc}
     \end{subfigure}
     \begin{subfigure}[h!]{0.47\textwidth}
         \centering
         \includegraphics[width=\textwidth,trim={46.1cm 13cm 19cm 12cm},clip]{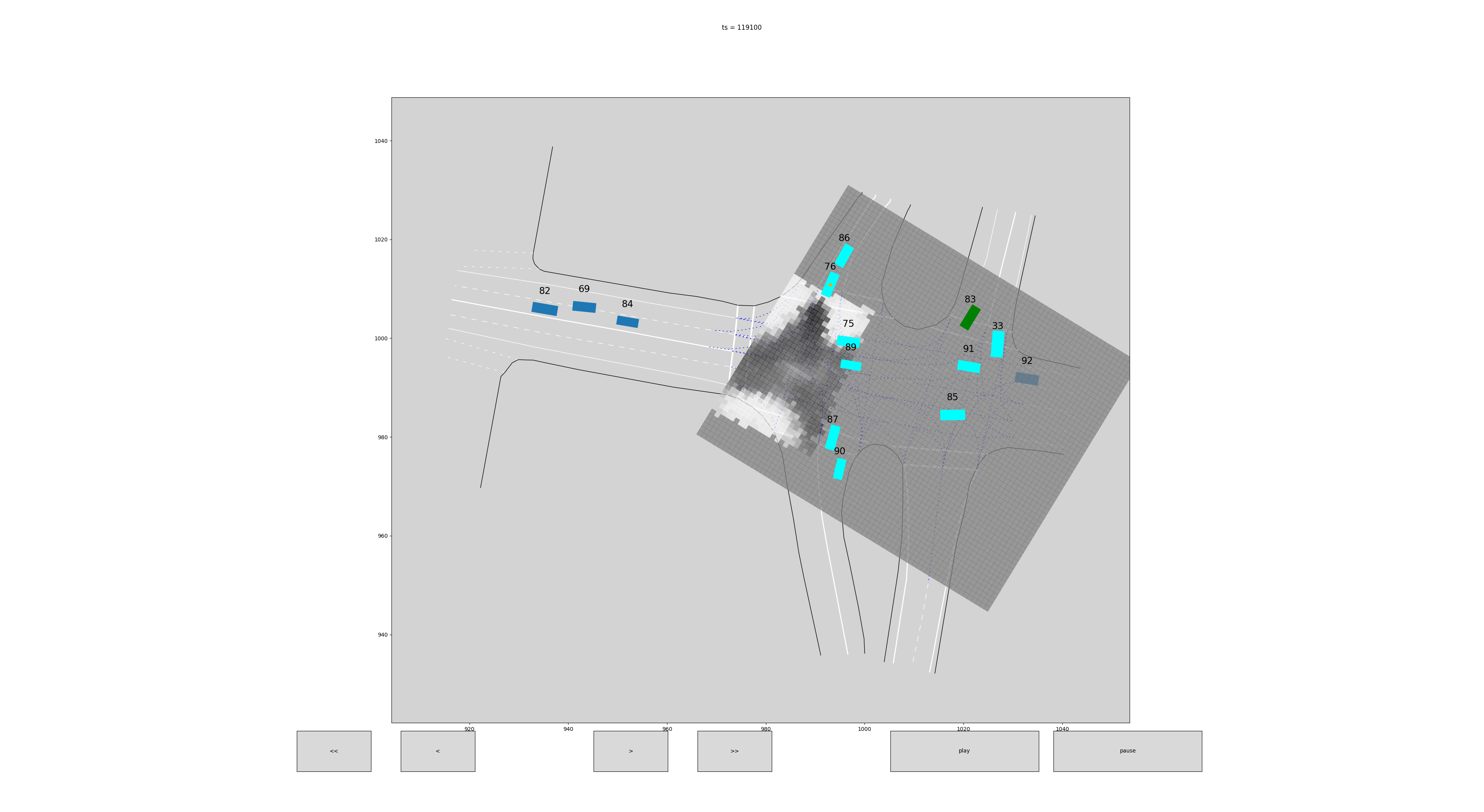}
         \caption{Vehicle 76: stopped.}
         \label{fig:vehicle_76}
     \end{subfigure}
        \caption{\small Individual inferred OGMs for some of the observed drivers for the scene in \cref{fig:stopped,fig:moving} using our proposed CVAE driver sensor model. The observed drivers are shown in cyan and their trajectories for the past \SI{1}{\second} ($s_{h}^{1:T}$) in orange. We focus on the observed drivers 33, 76 and 85. The inferred OGMs $\hat{\mathcal{M}}_{33}$, $\hat{\mathcal{M}}_{76}$, and $\hat{\mathcal{M}}_{85}$ are shown for their most likely latent mode within the ego vehicle's OGM, which is assumed to be fully occluded for easier visualization. The OGMs depict free (white), occluded (gray), and occupied (black) space. The OGMs in \cref{fig:vehicle_33_stopped,fig:vehicle_85,fig:vehicle_76} are from the same time step as \cref{fig:stopped}. The OGM in \cref{fig:vehicle_33_acc} is from the same time step as \cref{fig:moving}. The CVAE infers occupied space ahead for the stopped vehicles (\cref{fig:vehicle_33_stopped,fig:vehicle_76}). For the moving vehicle in \cref{fig:vehicle_33_acc}, the CVAE yields mostly free space. The constant speed vehicle 85 is estimated to have occupied space where traffic commonly exists. These qualitative results match our intuition for the occlusion inference task.
        }
        \label{fig:results_full_pipeline_only_one_vehicle}
\end{figure*}

In \cref{fig:results_full_pipeline_only_one_vehicle}, we illustrate the individual inferred OGMs for some of the observed drivers in the scene depicted in \cref{fig:stopped,fig:moving}. The inferred OGMs $\hat{\mathcal{M}}_{h}$ are shown within the ego vehicle's OGM, which, for easier visualization, is assumed to be entirely occluded. As expected, the CVAE driver sensor model infers the stopped vehicles (\cref{fig:vehicle_33_stopped,fig:vehicle_76}) to have occupied space ahead and to the left of them (where oncoming traffic is approaching from). Driver 85, who is moving at a constant speed, further adds to the occupied space output of driver 33 as traffic often exists in that lane in the dataset. As driver 33 starts moving (\cref{fig:vehicle_33_acc}), the space ahead and to the left of them is estimated to be free, signaling to the ego vehicle that it may be safe to turn right. The space to the right of driver 33 may be occupied with cars that have passed by, but these cars do not affect driver 33's left turn maneuver. These occlusion inference results align with our intuition for the task.

\begin{figure*}[h!]
     \centering
     \begin{subfigure}[h!]{0.41\textwidth}
         \centering
         \includegraphics[width=\textwidth,trim={46.1cm 13cm 19cm 12cm},clip]{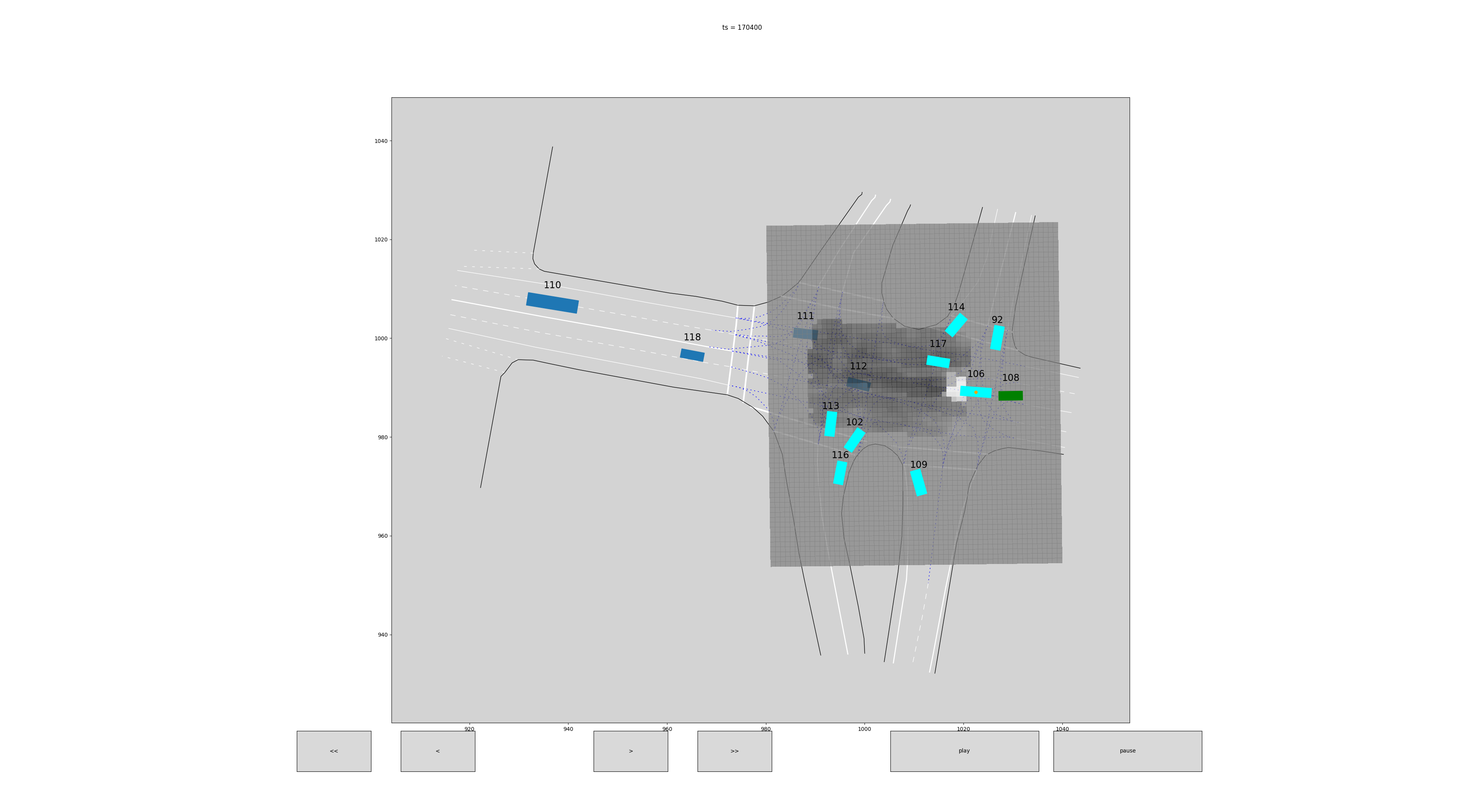}
         \caption{Vehicle 106: stopped.}
         \label{fig:vehicle_106_stopped_mode_1_only}
     \end{subfigure}
     \begin{subfigure}[h!]{0.41\textwidth}
         \centering
         \includegraphics[width=\textwidth,trim={46.1cm 13cm 19cm 12cm},clip]{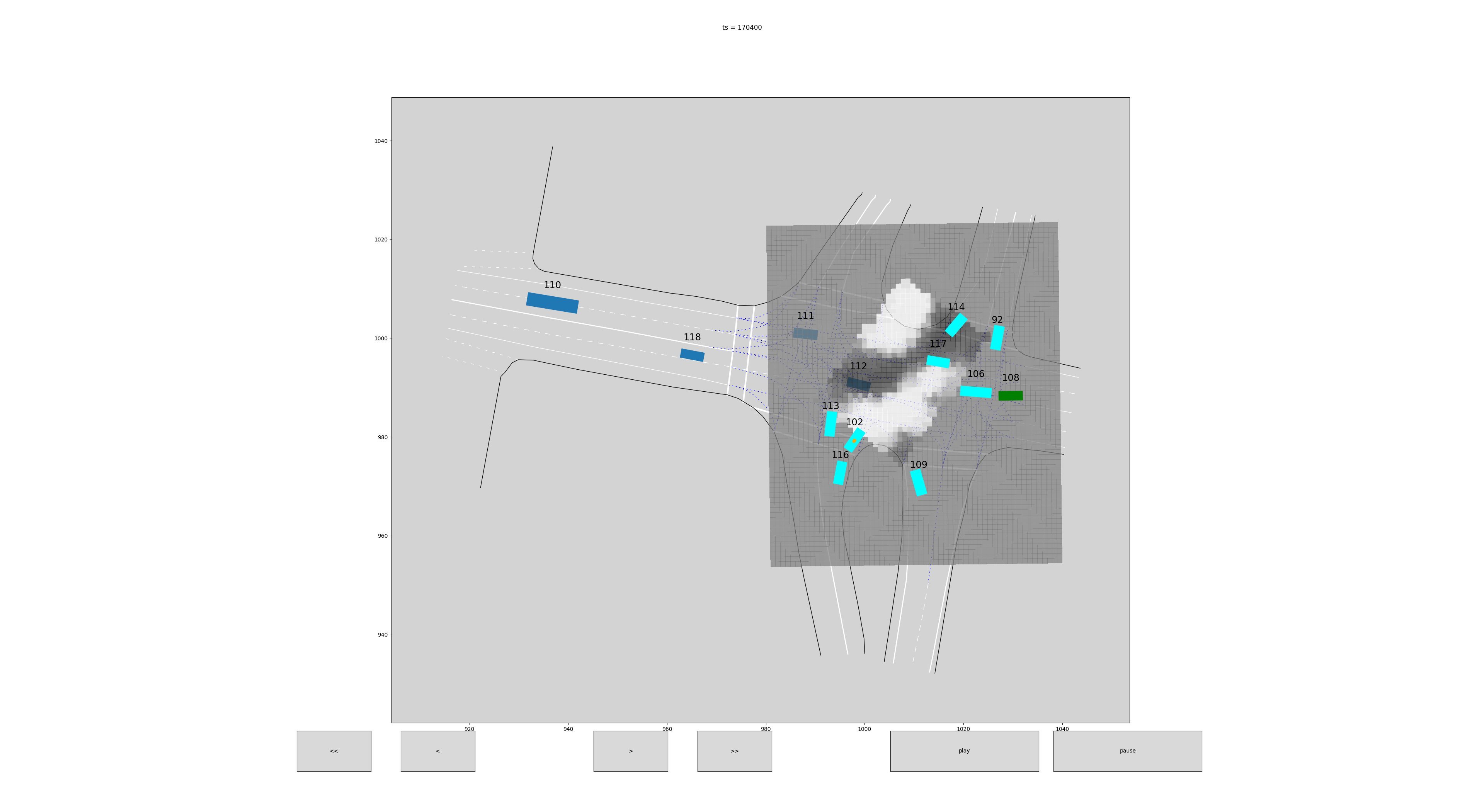}
         \caption{Vehicle 102: stopped.}
         \label{fig:vehicle_102_mode_1_only}
     \end{subfigure}
     \begin{subfigure}[h!]{0.41\textwidth}
         \centering
         \includegraphics[width=\textwidth,trim={46.1cm 13cm 19cm 12cm},clip]{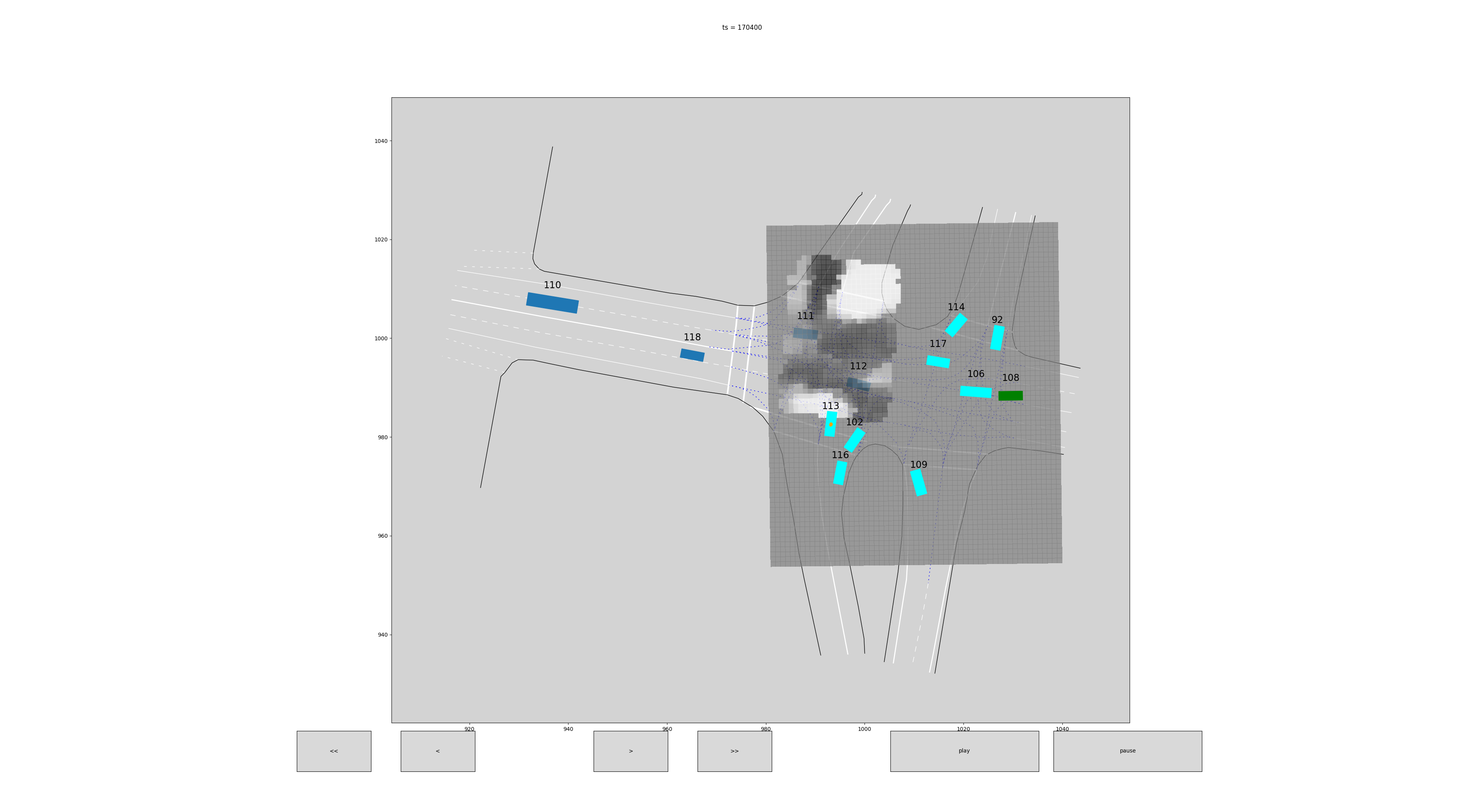}
         \caption{Vehicle 113: stopped.}
         \label{fig:vehicle_113_mode_1_only}
     \end{subfigure}
     \begin{subfigure}[h!]{0.41\textwidth}
         \centering
         \includegraphics[width=\textwidth,trim={46.1cm 13cm 19cm 12cm},clip]{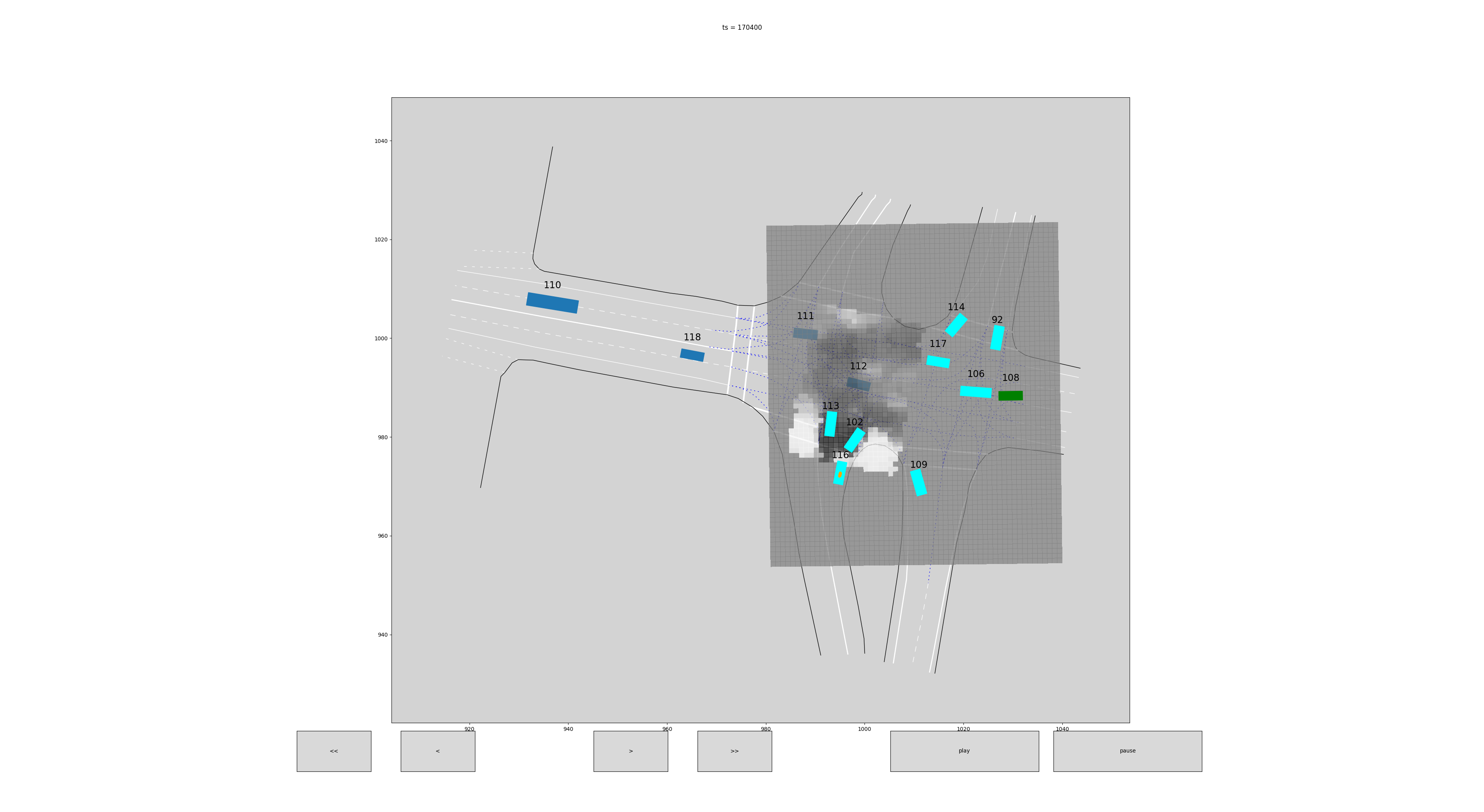}
         \caption{Vehicle 116: stopped.}
         \label{fig:vehicle_116_mode_1_only}
     \end{subfigure}
     \begin{subfigure}[h!]{0.41\textwidth}
         \centering
         \includegraphics[width=\textwidth,trim={46.1cm 13cm 19cm 12cm},clip]{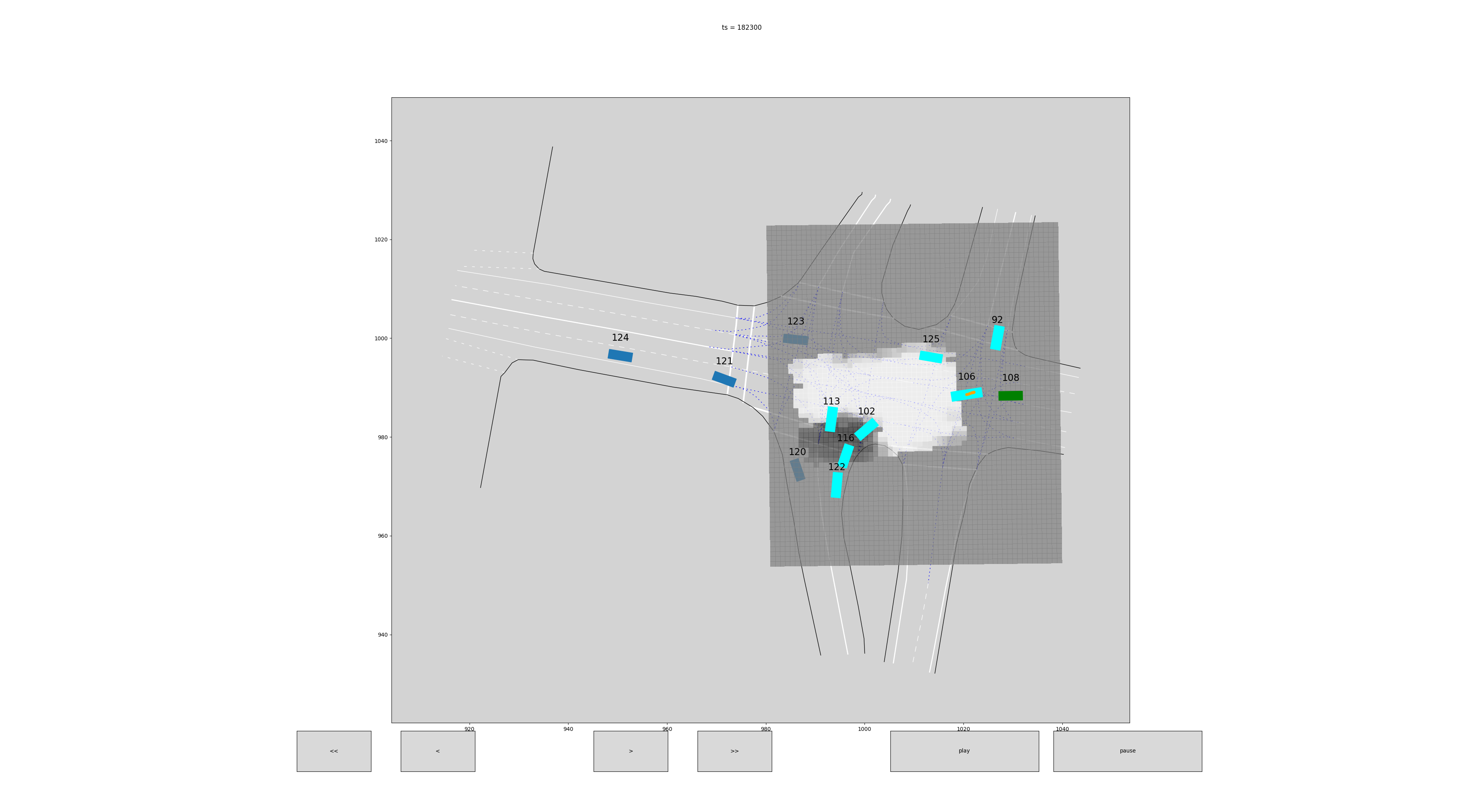}
         \caption{Vehicle 106: accelerating.}
         \label{fig:vehicle_106_moving_mode_1_only}
     \end{subfigure}
        \caption{\small Individual inferred OGMs for some of the observed drivers in the scene in \cref{fig:multiple_stopped,fig:multiple_accel} using our proposed CVAE driver sensor model. The observed drivers are shown in cyan and their trajectories for the past \SI{1}{\second} ($s_{h}^{1:T}$) in orange. We focus on the observed drivers 106, 102, 113, and 116. The inferred OGMs $\hat{\mathcal{M}}_{106}$ $\hat{\mathcal{M}}_{102}$, $\hat{\mathcal{M}}_{113}$, and $\hat{\mathcal{M}}_{116}$ are shown within the ego vehicle's OGM, which, for easier visibility, is assumed to be entirely occluded. The OGMs depict free (white), occluded (gray), and occupied (black) space. The CVAE driver sensor model successfully infers occupied space in the oncoming traffic lanes for the ego vehicle based on observed stopped vehicles, and free space in that region based on an observed accelerating vehicle.
        }
        \label{fig:results_full_pipeline2_only_one_vehicle}
\end{figure*}

Similarly, \cref{fig:results_full_pipeline2_only_one_vehicle} shows the individual OGM inferences for a subset of the observed drivers in the scene depicted in \cref{fig:multiple_stopped,fig:multiple_accel}. For the stopped drivers in \cref{fig:vehicle_106_stopped_mode_1_only,fig:vehicle_102_mode_1_only,fig:vehicle_113_mode_1_only,fig:vehicle_116_mode_1_only}, the CVAE successfully outputs occupied space in the inferred OGM, preventing these drivers from proceeding with their maneuvers. Notably, for driver 116 in \cref{fig:vehicle_116_mode_1_only}, who is stopped further away from the intersection, occupied space is also inferred for the drivers stopped ahead of them (in this case drivers 113 and 102). When driver 106 accelerates in \cref{fig:vehicle_106_moving_mode_1_only} to complete their left turn, the CVAE infers free space in the oncoming lane, as expected for them to complete their maneuver safely. 


\end{document}